\pdfoutput=1
\documentclass[11pt]{article}

\usepackage{ACL2023}

\usepackage{times}
\usepackage{latexsym}

\usepackage[T1]{fontenc}
\usepackage[utf8]{inputenc}

\usepackage{microtype}

\usepackage{inconsolata}

\usepackage{booktabs}
\usepackage{multirow}
\usepackage{graphicx}
\usepackage{amsmath}
\usepackage{caption}
\usepackage{colortbl}
\usepackage{pifont}

\title{Marathon: A Race Through the Realm of Long Context with\\Large Language Models}
\author{
Lei Zhang\textsuperscript{1,2} \quad
Yunshui Li\textsuperscript{1,2} \quad
Ziqiang Liu\textsuperscript{1,2} \quad
Jiaxi Yang\textsuperscript{1,2} \quad
Junhao Liu\textsuperscript{3} \\
\textbf{Longze Chen}\textsuperscript{1,2}  \quad
\textbf{Run Luo}\textsuperscript{1,2}  \quad
\textbf{Min Yang}\textsuperscript{1,2}\footnotemark[2] \\
\textsuperscript{1}Shenzhen Institute of Advanced Technology, Chinese Academy of Sciences\\
\textsuperscript{2}University of Chinese Academy of Sciences\\
\textsuperscript{3}University of California, Irvine \\
\texttt{\{lei.zhang2, min.yang\}@siat.ac.cn}
}

\begin{document}

\maketitle

\renewcommand{\thefootnote}{\fnsymbol{footnote}}
\footnotetext[2]{Min Yang is the corresponding author.}

\renewcommand{\thefootnote}{\arabic{footnote}}

\begin{abstract}
  With the advancement of large language models (LLMs) and the expansion of their context windows, existing long-context benchmarks fall short in effectively evaluating the models' comprehension and reasoning abilities in extended texts.
  Moreover, conventional benchmarks relying on F1 metrics often inaccurately score responses: they may undervalue correct answers that differ from the reference responses and overvalue incorrect ones that resemble the reference texts.
  In response to these limitations, we introduce \textbf{Marathon}, a novel evaluation benchmark that adopts a multiple-choice question format. It is specifically designed to overcome the constraints of previous benchmarks and provide a rapid, precise, and unbiased appraisal of the long-context comprehension skills of large language models.
  We conducted comprehensive evaluations on the Marathon benchmark with a range of state-of-the-art LLMs and assessed the effectiveness of various optimization strategies tailored for long-context generation.
  We anticipate that the Marathon benchmark and its associated leaderboard will enable a more precise and equitable evaluation of LLMs' capabilities in understanding and reasoning over extended contexts.
  \textbf{Marathon} is available at \url{https://github.com/Hambaobao/Marathon}.\footnote{We also provide an online evaluation website: \url{https://openbenchmark.online/marathon}}
\end{abstract}

\section{Introduction}

\begin{figure*}[!htbp]
  \centering
  \includegraphics[width=\textwidth]{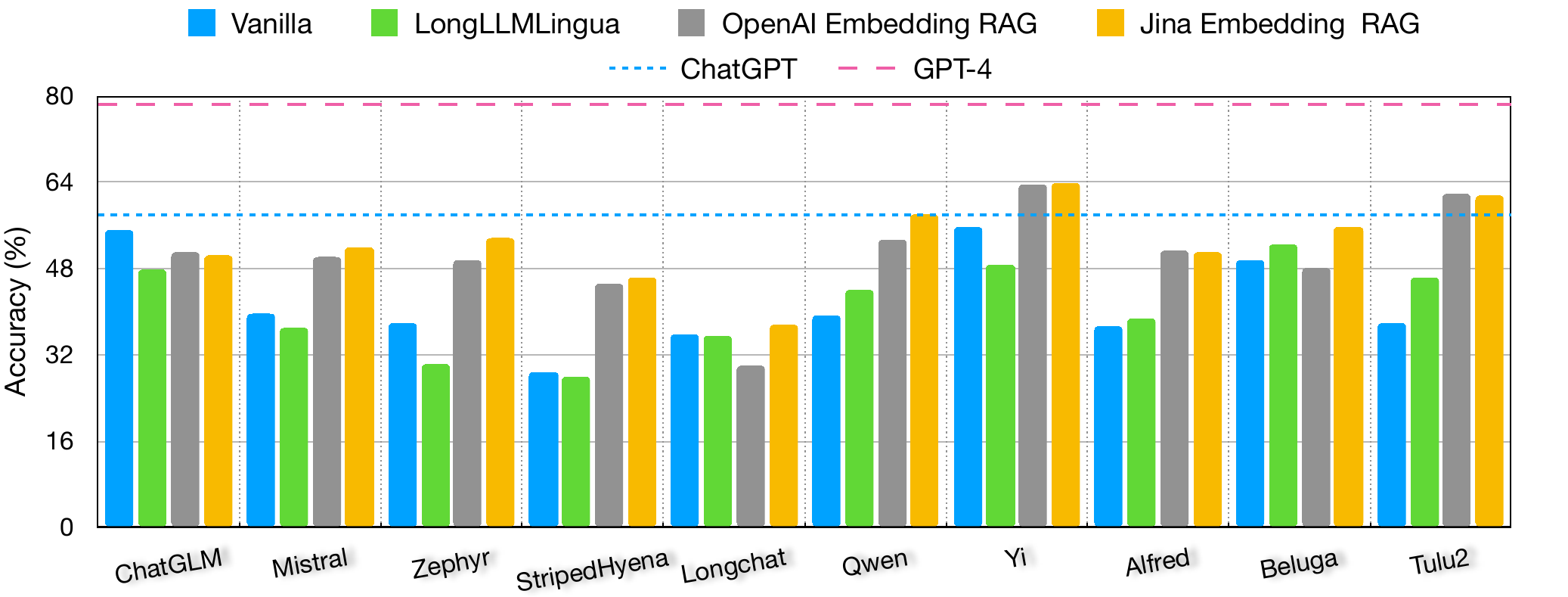}
  \caption{The overall accuracy of different models on Marathon.
    The x-axis represents the model, and the y-axis represents the average accuracy across all tasks.
    The different colors represent different methods of optimization.}
  \label{fig:overall_avg_accuracy}
\end{figure*}

In the rapidly evolving landscape of artificial intelligence technologies, the emergence of large language models (LLMs), as exemplified by ChatGPT \citep{openai2023gpt4}, showcases notable capabilities. The influence of these models extends beyond the well-established ChatGPT, gaining increasing prominence across diverse sectors.
Existing LLMs are typically built upon Transformer architectures, which demand memory and computational resources that grow quadratically with sequence length. Consequently, Transformer language models have historically been trained with relatively modest predetermined context windows.
%In the realm of LLMs, it is a common practice to train them with a predetermined context size. 
For instance, LLaMA \cite{touvron2023llama} employs a context size of 2048 tokens, while Llama2 \cite{touvron2023llama2} utilizes a context size of 4096 tokens. However, the pre-defined size imposes constraints on LLMs in various applications, such as summarizing extensive documents or addressing lengthy questions.
%LLMs are commonly trained with a predetermined context size, exemplified by 2048 tokens for LLaMA \cite{touvron2023llama} and 4096 tokens for LLaMA2 \cite{touvron2023llama2}.
%With the swift advancement of artificial intelligence technologies, large language models, exemplified by GPT-4 \citep{openai2023gpt4}, have exhibited remarkable competencies. Beyond the widely recognized ChatGPT, the utilization of large language models is progressively permeating diverse sectors.
%In the healthcare domain, a model named MedGPT \citep{kraljevic2021medgpt} aids individuals in seeking medical advice, while in the legal arena, LawGPT \citep{nguyen2023brief} offers assistance in legal inquiries, substantially diminishing the obstacles and expenses associated with accessing information.
%Similarly, in the field of software development, Github Copilot \citep{chen2021evaluating} enhances developer efficiency by expediting coding tasks, thereby significantly boosting productivity.

Significant research efforts have been devoted to extending the context length of LLMs.
Due to the prohibitive expense of training LLMs with extended context lengths from scratch, the predominant studies have endeavored to enhance the capabilities of LLMs to comprehend long contexts through fine-tuning.
These methods encompass extending the context window \cite{chen2023extending}, incorporating recurrent memory \cite{bulatov2024scaling}, employing sparse attention mechanisms \cite{xiao2023efficient}, and augmenting with external memory \cite{wang2023augmenting}.
Concurrently, an increasing multitude of benchmarks have been introduced to assess the long-context understanding capabilities of LLMs.
%Meanwhile, a growing number of benchmarks have been proposed to test the long-context understanding ability of LLMs.
%However, there has been relatively less attention devoted to evaluating long-context understanding abilities, where models are required to possess the ability to understand and reason over an extensive context. 
LongBench \cite{bai2023longbench} stands out as the first bilingual, multi-task benchmark specifically designed for the assessment of long-context understanding.
%This dataset still relies on the F1 score, which compares the model's response to a set of possible answers. However, the diverse nature of model-generated content means that these potential answers may not capture all correct responses, making the assessment of model performance less precise.
This dataset continues to depend on the F1 score, which evaluates the responses of LLMs against a predefined set of possible answers.
LooGLE \cite{li2023loogle}  encompasses intricate long dependency tasks, including event timeline reordering, comprehension/reasoning, and computation.
Nevertheless, the diverse nature of model-generated content introduces a challenge, as these predefined answers may not encompass all valid responses, thereby diminishing the precision of assessing model performance.
There is a growing demand for high-quality benchmarks characterized by significantly longer text lengths and more challenging tasks, ensuring comprehensive evaluations.

%Recently, there has been a growing effort to enhance the capabilities of LLMs in understanding long contexts. These methods encompass extending the context window \cite{}, incorporating recurrent memory \cite{}, employing sparse attention mechanisms \cite{}, and augmenting with external memory \cite{}. 

%Currently, advanced large language models are limited to responding to user queries within a narrow context window. This limitation arises from two primary factors. First, these models struggle to extract vital information from lengthy context. As \citet{liu_etal2023} outlines, these models efficiently identify key details only when they are near the start or end of the input. Retrieving important data from the middle becomes increasingly challenging as the text lengthens. Second, processing extended context demands greater computational power and time, leading to higher hardware requirements and a diminished user experience. Thus, improving the capability of these models to handle long context is essential, necessitating a thorough evaluation of their performance in this area. Current benchmarks for assessing performance on long context, such as LongBench \citep{bai2023longbench}, still rely on the F1 score, which compares the model's response to a set of possible answers. However, the diverse nature of model-generated content means that these potential answers may not capture all correct responses, making the assessment of model performance less precise.

In this study, we introduce a novel benchmark named \textbf{Marathon}, designed for long-context understanding and reasoning.
In particular, this benchmark is constructed upon the foundations established by LooGLE \citep{li2023loogle} and LongBench \citep{bai2023longbench}.
The contextual lengths within this benchmark span from 2K to over 260K characters. For each extensive context provided, an associated question is paired with four meticulously crafted response options.
These options have been carefully reviewed by humans and contain only one correct answer, with the remaining options designed to be highly misleading.
This design makes the Marathon benchmark a particularly challenging one.
The task for the large language model is to discern the accurate response option based on the extensive context provided.

%For every extensive context presented, an associated query is provided alongside four meticulously formulated response alternatives. The task for the large language model is to discern the accurate response option predicated on the extensive context supplied.

%In response to the aforementioned limitations, we introduces a novel benchmark for evaluating performance in extensive textual contexts, termed \textbf{Marathon}. This benchmark is constructed upon the foundations established by LooGLE \citep{li2023loogle} and LongBench \citep{bai2023longbench}. The contextual lengths within this benchmark span from 6K to in excess of 260K characters. For every extensive context presented, an associated query is provided alongside four meticulously formulated response alternatives. The task for the large language model is to discern the accurate response option predicated on the extensive context supplied.

% We conducted assessments on the \textbf{Marathon} benchmark using a spectrum of large language models, varying in parameter sizes and context window extents, alongside diverse long context compression techniques.
% As illustrated in Figure \ref{fig:overall_avg_accuracy}, our analysis revealed a general shortfall in these models' ability to process extended contexts efficiently.
% Furthermore, we identified that certain long context compression strategies could enhance the interpretative performance of large language models in specific scenarios.
The main contributions of this work are threefold:
\begin{itemize}
  \item We introduce a novel multiple-choice long context benchmark that comprehensively evaluates the long context understanding and reasoning capabilities across 10 leading open-source large language models, as well as ChatGPT and GPT-4, covering six diverse types of tasks.
  \item We compare two prevalent methods for long context optimization (Prompt Compression and Retrieval Augmented Generation) along with two leading embedding models, assessing their impact on enhancing the long context reasoning abilities of large language models.
  \item Our findings reveal a general tendency among current open-source large language models to generate lengthier responses, accompanied by a notable deficiency in following instructions accurately.
\end{itemize}

\section{Related Work}

\subsection{Prompt Compression}
Although larger context windows enable large language models to handle longer contextual information, processing long-context information requires a significant amount of computing resources and places high demands on hardware.
It also necessitates longer computational time, even in the inference stage.
Therefore, some methods like LLMLingua \citep{jiang_etal_2023_llmlingua} and LongLLMLingua \citep{jiang_etal_2023_longllmlingua} have been proposed to compress long contexts.

\subsection{Retrieval Augmented Generation}
Retrieval Augmented Generation (RAG) was originally proposed and applied to NLP tasks in \citep{Retrieval_Augmented_Generation}, and it has now become a mainstream method for improving the generation capability of large language models.
RAG can extract the most relevant data from external knowledge bases and hand it over to the large language model for processing.
This can alleviate the hallucination problem of large language models and enable people to trace the source of the content generated by large language models, ensuring the reliability of the generated content.
Additionally, RAG can also be used to extract information from long documents that is most relevant to the user's query.
This ensures that key information required to provide correct answers to questions is not lost while reducing the length of the context.
Many projects such as Longchain \footnote{\url{https://github.com/langchain-ai/langchain}} and LlamaIndex \footnote{\url{https://github.com/jerryjliu/llama_index}} have achieved significant progress in combining RAG with large language models, greatly facilitating related research in this direction.

\subsection{Long Context Models}
The ability of large language models for handling long contexts has become increasingly important.
ChatGPT \footnote{\url{https://chat.openai.com}} supports a window size of 16k, while GPT-4 supports a window size of 128k, and Claude-2.1 supports a window size of 200k\footnote{\url{https://www.anthropic.com/index/claude-2-1}}.
Many open-source large language models have started to expand the size of their context window.
Longchat \citep{longchat2023} and MPT \citep{MosaicML2023Introducing} have achieved a window size of 16k, while Mistral \citep{jiang2023mistral} and Zephyr \citep{tunstall2023zephyr} have achieved a window size of 32k.
By utilizing an adapted Rotary Embedding \citep{su2022roformer} and sliding window \citep{beltagy2020longformer} during fine-tuning, MistralLite, based on Mistral, has achieved a window size of 128k, enabling large language models to handle even longer contextual information.

\begin{figure*}[!h]
  \centering
  \includegraphics[width=\textwidth]{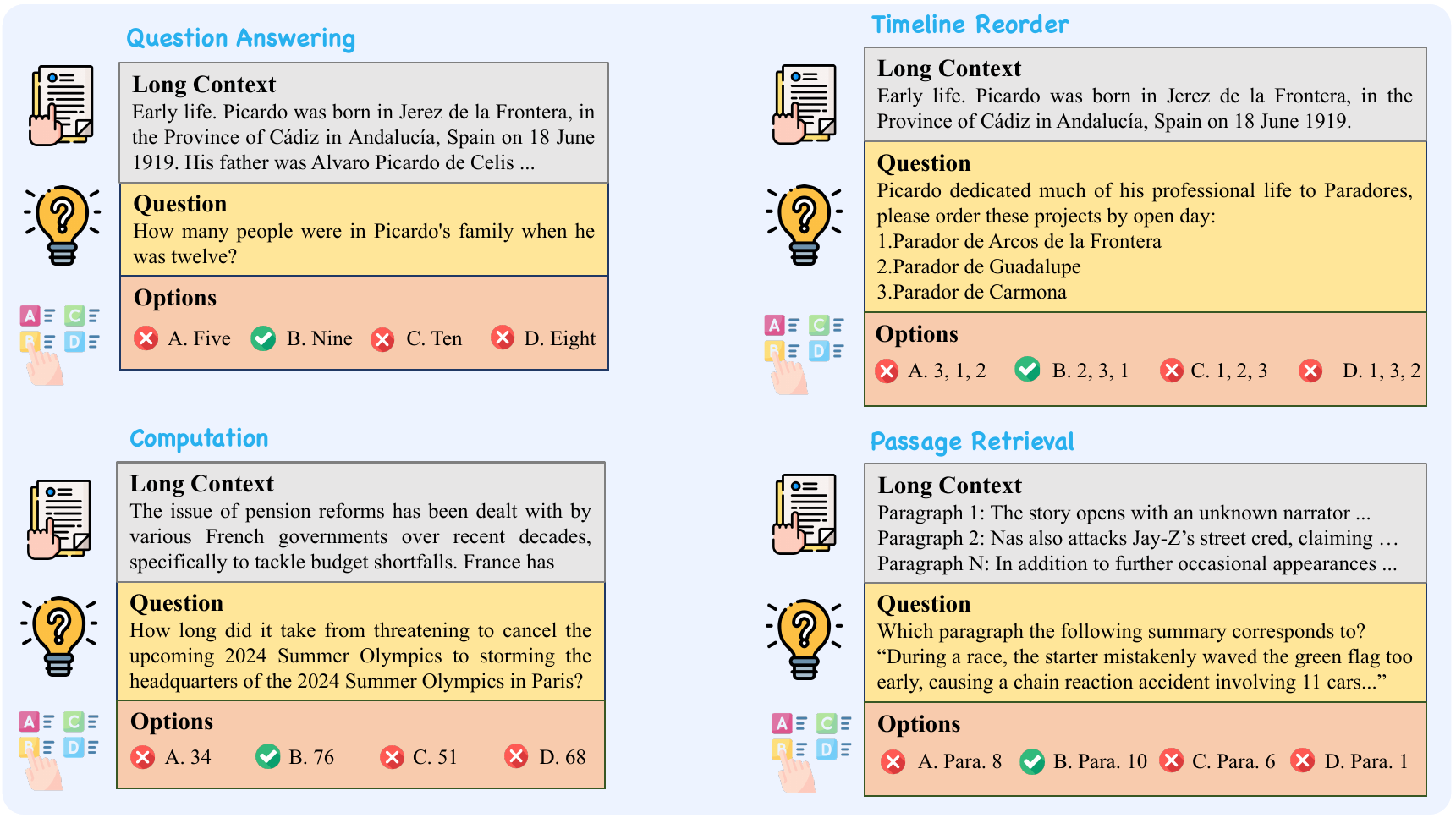}
  \caption{Examples of test cases in the benchmark. The context is truncated for display purposes.}
  \label{fig:example}
\end{figure*}

\subsection{Long Context Benchmarks}
There have been many recent benchmarks \citep{tay2020long,pang2022quality,shaham2022scrolls} used to assess the long context processing ability of large language models, such as LooGLE \citep{li2023loogle} and LongBench \citep{bai2023longbench}.
% LongBench includes two languages: Chinese and English, which utilizes data from HotpotQA \citep{yang2018hotpotqa},
% 2WikiMultihopQA \citep{ho_etal_2020_constructing},
% MuSiQue \citep{trivedi2022musique},
% DuReader \citep{he2018dureader},
% NarrativeQA \citep{kovcisky2018narrativeqa},
% Qasper \citep{dasigi2021dataset},
% GovReport \citep{huang2021efficient},
% QMSum \citep{zhong2021qmsum},
% MultiNews \citep{fabbri2019multi},
% VCSUM \citep{wu2023vcsum},
% TriviaQA \citep{joshi2017triviaqa},
% SAMSum \citep{gliwa2019samsum},
% LCC \citep{guo2023longcoder},
% and RepoBench-P \citep{liu2023repobench} to construct evaluation data for 14 English tasks, 5 Chinese tasks, and 2 code tasks.
\citet{liu_etal2023} on the other hand, noticed that the position of key information in long contexts greatly affects the capability of large language models to correctly understand and process text.
Therefore, they used the NaturalQA \citep{Natural_Questions} dataset to construct a new benchmark to test the impact of different positions of key information in long context on the text processing capability of large language models.

\citet{pang2022quality} proposed a multiple-choice long context benchmark, but its average text length is only 5k, which is significantly below the context window of mainstream large language models.
Although SCROLLS \citep{shaham2022scrolls}, LooGLE \citep{li2023loogle}, and LongBench \citep{bai2023longbench} have constructed a relatively comprehensive set of evaluation tasks, the evaluation metrics used are still F1-score, Bleu or Rouge, which cannot accurately evaluate the ability of large language models to handle and understand long contexts.

\section{Marathon}
Present benchmarks for evaluating large language models primarily use a multiple-choice format, highlighted by studies such as MMLU \citep{hendrycks2021measuring} and C-Eval \citep{huang2023ceval}.
This multiple-choice approach helps prevent situations where large language models produce correct answers but are scored low due to missing corresponding elements in the reference answers or when they produce incorrect answers that are scored high because they resemble the reference answers closely.
Therefore, influenced by LooGLE \citep{li2023loogle} and LongBench \citep{bai2023longbench}, we developed a multiple-choice, long-context benchmark to more accurately evaluate the ability of large language models to understand extended contexts.

\subsection{Overview}
The Marathon benchmark includes six tasks: \emph{Comprehension and Reasoning}, \emph{Multiple Information Retrieval}, \emph{Timeline Reorder}, \emph{Computation}, \emph{Passage Retrieval}, and \emph{Short Dependency QA}.
These tasks are grouped into four categories based on the type of questions they involve: Question Answering, Timeline Reordering, Computation, and Passage Retrieval.
Table \ref{tab:marathon_statistics} provides the number of test samples for each task.
Figure \ref{fig:example} presents example questions for each category.

\begin{table}[!h]
  \centering
  \resizebox{0.95\columnwidth}{!}{
    \begin{tabular}{c|c}
      \toprule
      \textbf{Task}                  & \textbf{No. Samples} \\
      \midrule
      Comprehension and Reasoning    & 357                  \\
      \midrule
      Multiple Information Retreival & 341                  \\
      \midrule
      Timeline Reorder               & 152                  \\
      \midrule
      Computation                    & 97                   \\
      \midrule
      Passage Retrieval              & 300                  \\
      \midrule
      Short Dependency QA            & 283                  \\
      \midrule
      \textbf{Total}                 & \textbf{1530}        \\
      \bottomrule
    \end{tabular}
  }
  \caption{
    Statistics of Marathon.
  }
  \label{tab:marathon_statistics}
\end{table}

\subsection{Construction}
All the test samples in the benchmark are in the form of multiple-choice questions, with each question containing one correct answer option and several distractor options.
We use GPT-4 to generate the distractor options for each question.
For each question, we divide the long context into multiple fragments of length 12,000 tokens and randomly select one fragment.
We require GPT-4 to generate three distractor options based on the given context fragment, question, and correct answer.
The purpose of this approach is to avoid using excessively long context that exceeds GPT-4's context window, which may affect the accuracy of the generated results.
By using shorter contexts, we can obtain distractor options that are more relevant to these shorter contexts.
The distractor options for all test samples were generated using the OpenAI \texttt{GPT-4-0613} API, with a total expenditure of approximately \$900.

Finally, to ensure the effectiveness and accuracy of these distractor options, we manually verify the options of each test sample.

\subsection{Question Answering}
\emph{Comprehension and Reasoning}, \emph{Multiple Information Retrieval} and \emph{Short Dependency QA} are all types of traditional question-answer formats.
The difference lies in the fact that \emph{Comprehension and Reasoning}, \emph{Multiple Information Retrieval} are selected from the Long Dependency QA dataset in LooGLE \citep{li2023loogle}, while \emph{Short Dependency QA} is selected from the Short Dependency QA dataset in LooGLE \citep{li2023loogle}.
In the question-answering tasks, each question is accompanied by a corresponding long context, and the large language model is required to infer the correct answer according to the long context.
For the \emph{Short Dependency QA} task, the relevant content for the correct answer is relatively concentrated within the long context.
For \emph{Comprehension and Reasoning} and \emph{Multiple Information Retrieval} tasks, the content relevant to the correct answer is more scattered throughout the long context.
Therefore, the large language model needs to possess strong long context understanding capability in order to solve the question correctly.

In the upper left of Figure \ref{fig:example}, an example of a Question Answering task is provided.
The question asks the large language model to answer a related question based on the content in the long context.
% The options include one correct option and three distractor options.

\begin{figure*}[!h]
  \centering
  \includegraphics[width=\textwidth]{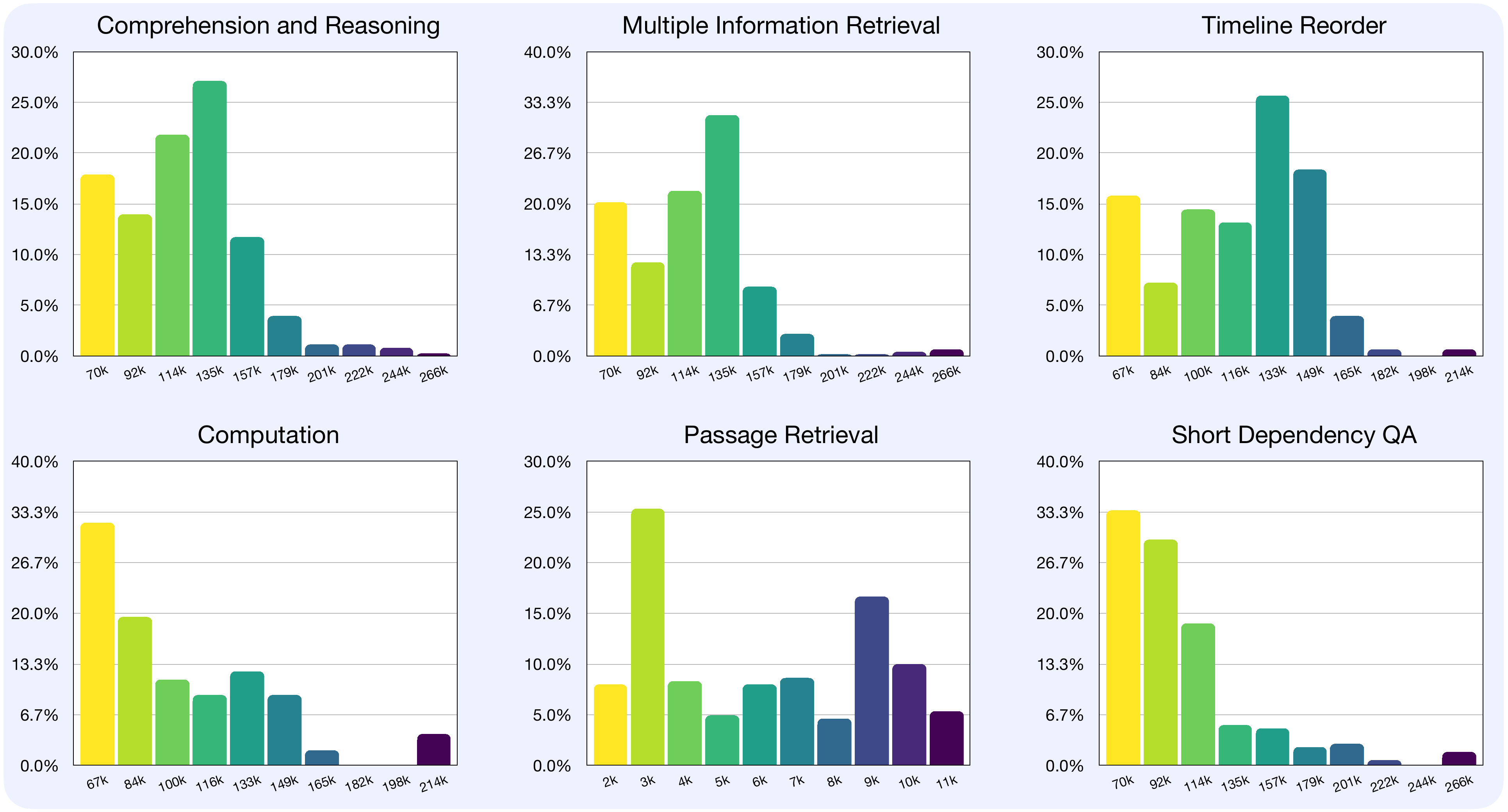}
  \caption{The distribution of context lengths for six tasks in the Marathon benchmark.}
  \label{fig:length}
\end{figure*}

\subsection{Timeline Reorder}
\emph{Timeline Reorder} task is a relatively novel question-answering task.
Unlike traditional question-answering tasks, in the \emph{Timeline Reorder} task, the question format requires large language models to sort a series of events described in a long context according to their chronological order.
This task aims to examine the large language models' understanding of temporal relationships.
Due to the dispersed distribution of events that need to be sorted by chronological order in the long context, large language models not only need to possess a correct understanding of temporal order but also require strong long context processing capabilities to answer correctly, which makes it a challenging task.

In the upper right of Figure \ref{fig:example}, an example of the \emph{Timeline Reorder} task is provided.
The question requires the large language model to sort three events mentioned in the long context according to their chronological order.
% The options include one correct order option and three incorrect order options.

\subsection{Computation}
\emph{Computation} task is also different from traditional question-answering tasks.
Its question format involves providing a question related to numerical computation and requires the large language model to perform numerical calculations based on relevant content in the long context.
For example, it may require calculating the number of children a certain character has at a specific time point, considering that the long context describes the character's life events, including the death of a child due to illness, which may affect the number of the character's offspring at subsequent time points.
Therefore, to answer this question correctly, the large language model not only needs to be able to perform ordinary numerical calculations but also needs to capture all the key information related to the question.
Compared to traditional computation and question answering tasks, this task is more challenging and can better reflect the large language model's capability to comprehend long context.

In the bottom left of Figure \ref{fig:example}, an example of a \emph{Computation} task is provided.
The question requires the large language model to complete a numerical calculation question based on the content in the long context.
% The options include one correct answer option and three incorrect answer options.

\subsection{Passage Retrieval}
\emph{Passage Retrieval} task is one form of task in the LongBench \citep{bai2023longbench}.
In order to enhance the diversity of our benchmark tasks, we have sampled 300 test data from the Passage Retrieval task in LongBench \citep{bai2023longbench}, and reformed them into multiple-choice format using the method mentioned above.
We have incorporated this task into our benchmark.
The \emph{Passage Retrieval} task requires large language models to locate the paragraph in a long context that corresponds to the given description in the question.
Since the test data of the \emph{Passage Retrieval} task is sampled from LongBench \citep{bai2023longbench}, there are some limitations in terms of context length and timeliness.
However, it remains a highly valuable task format.
In future work, we will update its content to make it more suitable for the current needs of evaluating large language models.

A sample of the \emph{Passage Retrieval} task is provided in the bottom right of Figure \ref{fig:example}.
The task requires large language models to locate the paragraph in a long context that corresponds to the given description in the question.
% The options consist of one correct order option and three incorrect order options.

\begin{figure*}[!h]
  \centering
  \includegraphics[width=0.98\textwidth]{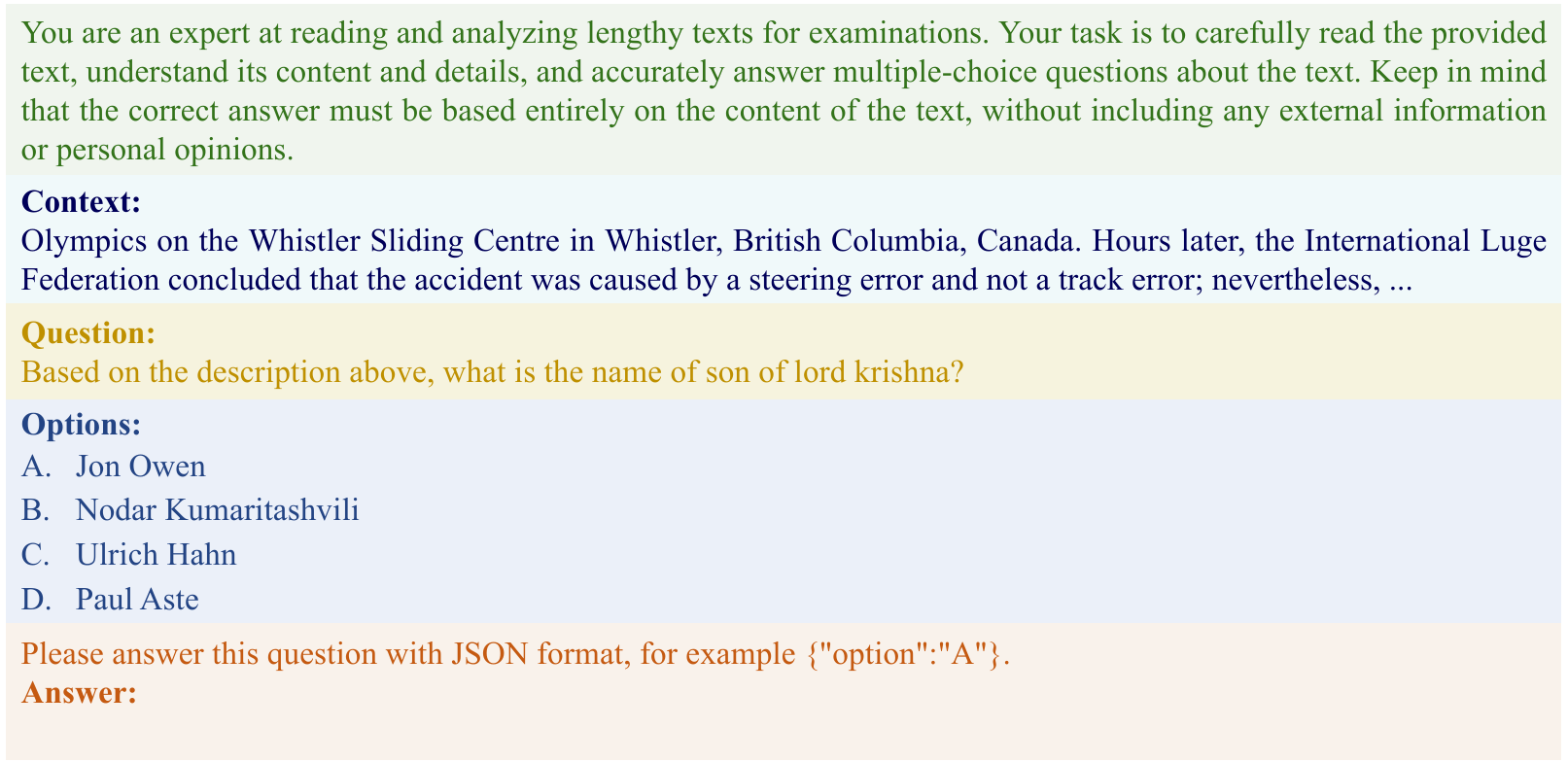}
  \caption{An example of a test prompt. The context is truncated for display purposes.}
  \label{fig:prompt}
\end{figure*}

\section{Experiments}

\subsection{Experimental Setup}

\subsubsection{Models}
In this analysis, we incorporated a diverse array of models, distinguished by their parameter sizes, which span from 7B to 70B, and their context window capacities, extending from 8K to 200K.
Additionally, the evaluation encompassed models constructed on a state-space architectural framework.
The models scrutinized in this investigation comprises ChatGLM3-6B-32K \citep{zeng2022glm,du2022glm}, Mistral-7B-Instruct-v0.1 \citep{jiang2023mistral}, Zephyr-7B-$\beta$ \citep{tunstall2023zephyr}, StripedHyena-Nous-7B \citep{stripedhyena}, Longchat-13B-16K \citep{longchat2023}, Qwen-14B-Chat \citep{qwen}, Yi-34B \citep{yi2023}, Alfred-40B-1023 \citep{alfred-40b-1023}, StableBeluga-2-70B \citep{StableBelugaModels}, Tulu-2-DPO-70B \citep{ivison2023camels}, ChatGPT-1106 \footnote{\url{https://chat.openai.com}}, and GPT-4-1106-preview \citep{openai2023gpt4}.

\subsubsection{Prompt Template}
Figure \ref{fig:prompt} illustrates the prompt used in our model evaluation process.
The system prompt is denoted in green, the provided long context in cyan, the question related to the long context in yellow, the quartet of options in blue, and the orange segment delineates the response format for the model, accompanied by a concrete example.
The instructions and response templates may vary across different models.
To ensure consistency, we adjust the prompts during our evaluations to match the templates used during the models' training phases.

\subsubsection{Optimization Methods}
In this evaluation, we first assessed the inherent ability of various models to comprehend long contexts.
Then, we evaluated the current mainstream methods for handling long contexts: Compression and RAG.
Specifically, for the compression method, we assessed LongLLMLingua \citep{jiang_etal_2023_longllmlingua}, while for the RAG method, we evaluated two retrieval approaches, one based on OpenAI Embedding and the other on Jina Embedding \citep{gunther2023jina}.

\subsubsection{Evaluation Metric}
To assess the long-context comprehension abilities of large language models, we employed a multiple-choice format for the questions, each designed with several options, among which only one is the correct answer.
\begin{equation*}
  \text{Accuracy} = \frac{\text{No. of Correct Answers}}{\text{No. of Total Answers}}
\end{equation*}

\subsubsection{Post Processing}
As illustrated in the prompt template of Figure \ref{fig:prompt}, we require the model to output responses in JSON format.
However, current large language models are not capable of following user instructions 100\% of the time.
Therefore, after the model generates a response, we first parse the output using JSON format.
For outputs that do not comply with the JSON format, we use regular expressions to extract the model's responses.

\subsection{Implementation Details}

\subsection{Vanilla}
For models evaluated using the Vanilla method, we adopt the same strategy as Longbench, controlling the length of the input sequence within the context window range of the model to be evaluated by deleting the content in the middle.

\subsubsection{LongLLMLingua}
For LongLLMLingua, we set the \emph{compression rate} to 0.5,
the \emph{dynamic context compression ratio} to 0.4,
the \emph{condition in question} to \texttt{"after"},
and the \emph{condition compare} to \texttt{True}.
We also sort the compressed contexts based on their importance.

\subsubsection{Embedding RAG}
For Embedding RAG, we utilize the ServiceContext and VectorStoreIndex of the Llama-Index \footnote{\url{https://github.com/jerryjliu/llama_index}}.
We employ various models as LLMs (Language Models), testing the OpenAI Embedding model and the Jina Embedding model as Embedding Models, respectively.
The default parameter settings are retained, with a chunk size of 1024 and a top-k value of 2.
As for Jina Embedding, we set the pooling method to "mean" to align with Jina's encode implementation.

\subsubsection{Hardware}
All experiments in this evaluation were conducted on a server with 4 * A100 80GB GPUs.
We set the batch size to 1. For models with a scale of 7B and 13B, we use one GPU; for models around 30B, we use two GPUs; and for models at the 70B scale, we use four GPUs.

\begin{figure*}[!htbp]
  \centering
  \includegraphics[width=\textwidth]{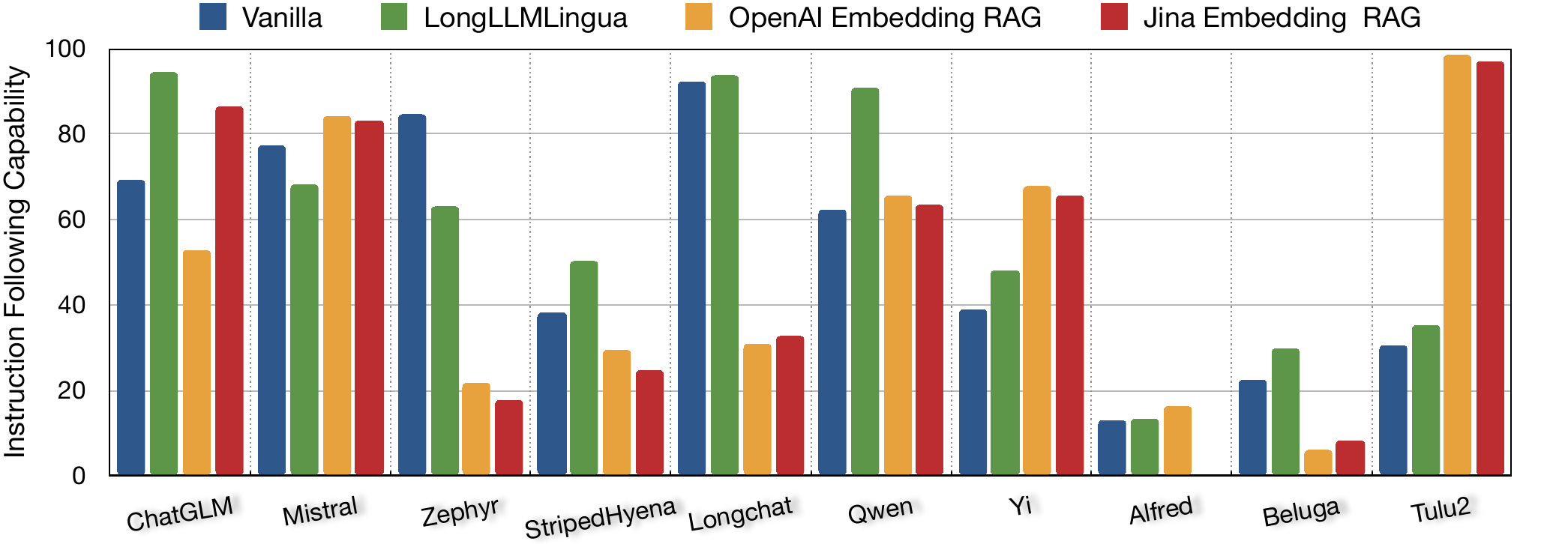}
  \caption{The instruction following the capability of different models.
    The x-axis represents the model, and the y-axis represents the instruction following capability.
    The different colors represent different methods of optimization.}
  \label{fig:Instruction_following}
\end{figure*}

\subsection{Results}
\subsubsection{Main results}
The overall accuracy of various models on the Marathon benchmark is depicted in Figure \ref{fig:overall_avg_accuracy}.
Detailed performance metrics of these models, utilizing distinct optimization techniques across a range of tasks, are presented in Table \ref{Super_Marathon_statistics} within the appendix.
To facilitate a more comprehensive comparative analysis of the outcomes, Figures \ref{fig:Visual_Comprehension_and_Reasoning}, \ref{fig:Visual_Multiple_Information_Retrieval}, \ref{fig:Visual_Timeline_Reorder}, \ref{fig:Visual_Computation}, \ref{fig:Visual_Passage_Retrieval}, and \ref{fig:Visual_Short_Dependency_Question_Answering} are provided in the appendix.
The analysis indicates that the OpenAI Embedding Retrieval and Jina Embedding Retrieval models exhibit superior performance relative to the LongLLMLingua compression.

Moreover, all examined models exhibit diminished accuracy on both the \emph{Timeline Reorder} and \emph{Computation} tasks relative to their performance on alternative tasks.
The implementation of the LongLLMLingua failed to yield any notable enhancements, and the advancements facilitated by the RAG were similarly constrained.

\subsubsection{Vanilla}
Within the subset of the Vanilla method, the Yi-34B model, characterized by its 34 billion parameters, attains the highest accuracy, registering at 55.91\%.
This is closely followed by the ChatGLM3-6B-32K, which, despite its more modest parameter count of 6B, achieves an accuracy of 55.05\%.
Subsequently, the Beluga-70B model, notable for its context window limitation of 4K tokens, records an accuracy of 49.51\%.
The average accuracy observed across the remaining models shows minimal variance, with none exceeding the 40\% threshold.

\subsubsection{LongLLMLingua}
In contrast to the Vanilla approach, the implementation of LongLLMLingua yielded marginal improvements in accuracy for certain models: Qwen witnessed an enhancement of 4.85\%, Alfred experienced a 1.51\% increase, Beluga saw a 3.08\% uplift, and Tulu2 benefited from an 8.64\% augmentation.
Conversely, this methodology had a detrimental effect on the performance of other models: ChatGLM3 encountered a 7.14\% decrement in accuracy, Mistral suffered a 2.8\% reduction, Zephyr experienced a significant 7.74\% decrease, StripedHyena and Longchat showed a marginal decline of 0.10\% and 0.26\% respectively, and Yi's accuracy diminished by 7.25\%.

\subsubsection{OpenAI Embedding RAG}
When juxtaposed with the baseline Vanilla methodology, the incorporation of OpenAI Embedding Retrieval notably enhances accuracy for several models: Mistral's accuracy improved by 10.37\%, Zephyr's by 11.66\%, StripedHyena's by 16.26\%, Qwen's by 14.19\%, Yi's by 7.65\%, Alfred's by 14.05\%, and Tulu2's by an impressive 24.05\%.
Conversely, this approach has been observed to negatively impact the accuracy of certain models, with ChatGLM3 experiencing a 4.06\% reduction, Longchat a 5.92\% decrease, and Beluga a slight decline of 1.27\%.

\subsubsection{Jina Embedding RAG}
Relative to the foundational Vanilla approach, the adoption of Jina Embedding Retrieval has led to accuracy enhancements across a majority of the evaluated models.
Notably, Mistral's accuracy experienced a 12.23\% increase, Zephyr's accuracy rose by 15.82\%, StripedHyena's accuracy increased 17.37\%, Longchat saw a 1.91\% improvement, Qwen's accuracy was augmented by 18.15\%, Yi's accuracy escalated by 7.9\%, Alfred's accuracy advanced by 13.93\%, Beluga's accuracy grew by 6.21\%, and Tulu2's accuracy surged by 23.60\%.

\subsection{Instruction Following Capability}
In our evaluation, numerous models exhibited limited ability to follow instructions accurately.
We explicitly requested responses in JSON format, as exemplified by a sample provided.
Nonetheless, models occasionally responded in alternate formats or attempted JSON responses that were either incomplete or incorrect.
Our statistical analysis, summarized in Table \ref{JSON_statistics}, categorizes responses as "JSON" for correct JSON format, "JSON-like" for flawed attempts at JSON due to errors like truncation or formatting issues, and "Plain Text" for responses in other formats.
For a clearer comparison of models' ability to follow instructions after applying various optimizations, we focused on the rate of correct JSON responses as a measure of this capability, as depicted in Figure \ref{fig:Instruction_following}.

While Yi exhibited high accuracy in question answering, its compliance with instructions was notably lower, at 38.95\%.
In contrast, Beluga's adherence rate to instructions was even lower, at 22.48\%, despite its capabilities.
On the other hand, Longchat, despite its modest accuracy in answering questions, showcased a remarkable proficiency in following instructions, achieving a 92.29\% compliance rate, closely trailing behind ChatGPT's 99.61\%.
The three distinct optimization techniques in our assessment demonstrated efficacy in diminishing the context length.
However, it is noteworthy that none of these strategies consistently enhance the models' ability to follow instructions.

\section{Discussion}
\subsection{Tendency of Long Responses}
During our analysis, we observed that open-source large language models often generate lengthy responses, even with clear instructions for concise JSON-formatted answers.
This tendency results in the generation of extraneous content, necessitating post-processing to isolate the needed information.
Table \ref{JSON_statistics} presents statistics on the models' output formats, highlighting their instruction-following capabilities.
This issue likely stems from the models' training on predominantly long responses, making it challenging for them to comply with requests for brevity.

\subsection{State Space Models}
Recent studies, such as Mamba \citep{gu2023mamba}, highlight the advantages of state space models (SSMs) for long context reasoning tasks.
StripedHyena \citep{stripedhyena} innovatively merges SSMs with transformer structures, indicating a new direction in large language models.
Despite these advancements, our analysis reveals that StripedHyena underperforms in detailed long context question answering compared to traditional transformers and does not reduce memory usage effectively, even with advanced attention mechanisms like Flash Attention 2 \cite{dao2023flashattention2}.
These findings suggest the need for further optimization in State Space Models.

\subsection{JSON Format}
During the recent OpenAI Developer Day\footnote{\url{https://devday.openai.com}}, significant advancements in the capabilities of GPT-4 \citep{openai2023gpt4} were unveiled by OpenAI, notably the introduction of parallel function invocation and the specification of response formats in JSON.
The parallel function invocation allows for the concurrent execution of multiple utility functions by large language models, thereby facilitating the efficient completion of complex user tasks.
Moreover, the integration of JSON format for responses is instrumental in ensuring the seamless transmission of parameters and retrieval of results during function invocation, which is critical for the interoperability and functionality of AGI systems.

\section{Future Work}
\subsection{Document as Context}
Following the enhancements introduced at OpenAI Developer Day, GPT-4 \citep{openai2023gpt4} has been equipped with a \emph{Knowledge Retrieval} feature.
This allows the model to utilize user-uploaded documents for answering queries, marking a significant development in Retrieval-Augmented Generation (RAG) applications.
This trend suggests that future large language models will likely adopt similar functionalities, impacting the evaluation methodologies for long-context question answering.
Instead of embedding lengthy contexts into prompts, future benchmarks should focus on the models' ability to extract and utilize information from user-provided documents to respond to queries.
This approach necessitates a reevaluation of current benchmarks to align with these emerging capabilities.

\subsection{Multi-modal Long Context}
Models such as GPT4V \citep{openai2023gpt4} and Gemini \citep{geminiteam2023gemini} have exhibited robust capabilities in facilitating interactions that span both visual and linguistic modalities.
Likewise, open-source counterparts, including LLaVA \citep{liu2023visual} and MiniGPT-4 \citep{zhu2023minigpt4}, have demonstrated commendable performance in assessments tailored to multimodal contexts.
The utility of such models extends to various real-world applications that necessitate the processing of multimodal, extensive contexts, exemplified by the comprehensive analysis and synthesis of corporate annual reports.
These applications demand not only the capacity of large language models to comprehend and infer within long textual contexts but also necessitate the integration of visual understanding abilities.
Presently, the open-source community is lack of benchmarks specifically designed to evaluate the proficiency of models in handling extended, multimodal contexts.
Therefore, establishing a comprehensive benchmark for multimodal, long-context capabilities is of significant importance.

% \subsection{Evolving Online Benchmarks}
% The rapid advancement of large language models calls for evolving evaluation methods and benchmarks.
% Traditional static benchmarks, often compromised over time by data leakage and integration into training, become less effective for accurate assessments.
% Moreover, the development of benchmarks by isolated teams or researchers is not only inefficient but also faces challenges in continuously updating evaluation data. The solution lies in dynamic online benchmarks, which would draw on the collective expertise and resources of the open-source community to ensure a constantly updated repository of new tasks and evaluation methods.
% This model aims to keep pace with the fast-evolving capabilities of language models, offering a more effective and scalable assessment framework.

\section{Conclusion}
In this paper, we compared 10 open-source large language models, including variations in their parameter sizes and context windows, along with OpenAI's ChatGPT and GPT-4.
We assessed two prevalent optimization techniques, such as LongLLMLingua and RAG.
The experimental results indicate that RAG-based optimization enhances the performance of large language models within long-context scenarios for QA-type tasks. However, the improvement is limited for tasks involving Timeline Reorder and Computation.
Despite high accuracy in question-answering, these models show limited ability to follow instructions.

\section*{Acknowledgments}
Min Yang was supported by National Key Research and Development Program of China (2022YFF0902100), National Natural Science Foundation of China (62376262), the Natural Science Foundation of Guangdong Province of China (2024A1515030166), Shenzhen Science and Technology Innovation Program (KQTD20190929172835662), Shenzhen Basic Research Foundation (JCYJ20210324115614039).

\section*{Limitations}
\paragraph{Context Length Distribution.}
As depicted in Figure \ref{fig:length}, the distribution of context lengths within the Marathon benchmark exhibits a lack of uniformity.
The test instances corresponding to the tasks of \emph{Comprehension and Reasoning}, \emph{Multiple Information Retrieval}, \emph{Computation}, \emph{Short Dependency QA}, and \emph{Timeline Reorder} predominantly feature context lengths that are concentrated at, or below, 130K characters.
Conversely, test instances with context lengths surpassing 200K characters are notably scarce.

The test instances for the \emph{Passage Retrieval} task derive from the LongBench \citep{bai2023longbench} dataset, which accounts for the markedly shorter context lengths in comparison to those associated with the remaining five tasks.
This discrepancy underlies the superior performance metrics achieved by all models on the \emph{Passage Retrieval} task.
It is our intention to revise the test instances for \emph{Passage Retrieval} to ensure consistency in context lengths with the other tasks.
Furthermore, our ongoing efforts are directed towards augmenting the test instances for the remaining tasks, with the objective of achieving a uniform distribution of context lengths ranging from 60K to 260K characters across all tasks.

\paragraph{Evaluation.}
This paper presents a preliminary evaluation of optimization techniques for long contexts, which is not all-encompassing.
In terms of optimization strategies, our evaluation of the Retrieval-Augmented Generation (RAG) method was limited to the employment of the OpenAI and Jina Embedding systems, exemplifying leading commercial and open-source embedding models, respectively.
However, constraints related to time and financial resources precluded the examination of several advanced embedding systems, such as Voyage \citep{voyage_embedding}, Cohere \citep{cohere_embedding}, and BGE Embeddings \citep{bge_embedding}.
In the case of the Prompt Compression approach, aside from LongLLMLingua, there are other techniques like MemWalker \citep{chen2023walking} that merit future exploration to fully assess the advantages and drawbacks of each embedding model and optimization method.

Moreover, in scenarios involving long context, while model accuracy and adherence are crucial, the speed of inference and memory demand are also vital factors to consider.
This area features a variety of sophisticated optimization methods, including H2O \citep{zhang2023h2o} and StreamingLLM \cite{xiao2023efficient}.
Subsequent research will focus on evaluating the performance of these inference optimization methods in scenarios with extensive textual content, with an emphasis on their speed of inference, memory consumption, QA precision, and instruction following capability.

\section*{Ethical Considerations}
\paragraph{Data Source and Use.}
The benchmark leverages datasets that are publicly available and designated for research purposes.
We have ensured that the use of these datasets adheres to their respective licenses and terms of use, emphasizing that our utilization is strictly confined to academic and research contexts.

\paragraph{Content Sensitivity and Bias.}
Our benchmark has been meticulously curated to exclude any content that could be deemed sensitive, such as violence, discriminatory language, or adult material.

\paragraph{Transparency and Reproducibility.}
In the spirit of fostering an open and fair research community, we will make the questions, contexts, and options of our benchmark's test cases publicly available.
However, to maintain the integrity of the evaluation process, the correct answers to the test cases will not be disclosed.
Instead, we will provide an online evaluation platform where researchers can submit their models' responses for assessment.
This system is designed to ensure fairness and objectivity in the benchmarking process, allowing for an equitable comparison of different models' capabilities.

\bibliography{acl2023}
\bibliographystyle{acl_natbib}

\onecolumn
\appendix

\section{Dataset Construction}
\subsection{Data Collection}
The data in the Marathon benchmark primarily originates from LongBench \citep{bai2023longbench} and LooGLE \citep{li2023loogle}.
Table \ref{tab:Marathon_Sources} provides specific details regarding the sources of the data and the licensing information for the benchmark.

\begin{table*}
  \centering
  % \resizebox{!}{\textwidth}{
  \begin{tabular}{l|c|cccccccc}
    \toprule
    \textbf{Original Dataset}          & \textbf{License} & \textbf{C\&R} & \textbf{MIR} & \textbf{TR} & \textbf{Computation} & \textbf{PR} & \textbf{SDQA} \\
    \midrule
    LongBench \citep{bai2023longbench} & MIT              & \ding{51}     & \ding{51}    & \ding{51}   & \ding{51}            & \ding{55}   & \ding{51}     \\
    LooGLE \citep{li2023loogle}        & MIT              & \ding{55}     & \ding{55}    & \ding{55}   & \ding{55}            & \ding{51}   & \ding{55}     \\
    \bottomrule
  \end{tabular}
  % }
  \caption{Sources of data for six tasks in Marathon benchmark.}
  \label{tab:Marathon_Sources}
\end{table*}

\subsection{Data Processing}
For \emph{Comprehension and Reasoning}, \emph{Multiple Information Retrieval}, \emph{Short Dependency QA}, \emph{Timeline Reorder} and \emph{Computation} tasks, we use data from LooGLE \citep{li2023loogle} to construct the test samples.
For each question, we divide its corresponding long context into text segments of 12,000 tokens each and randomly select one segment to generate a distractor option, each segment is used no more than once.
We send the selected long context segment $C_i$, the question $Q$, and the correct answer $A$ in a specific format to GPT-4, requesting it to provide a distractor option $O_i$ based on $C_i$, $Q$, and $A$.

For \emph{Passage Retrieval} task, we construct test samples with Python program.
We sampled data from Longbench \citep{bai2023longbench} and increased the number of passages to extend the context length.

To ensure the effectiveness and accuracy of these distractor options, we manually verify the options of each test sample.

\section{Model Description}
\paragraph{ChatGLM.}
The latest open-source model in the ChatGLM series, ChatGLM3-6B, as outlined by \cite{zeng2022glm,du2022glm}, maintains several outstanding features of its predecessors, such as smooth dialogue capabilities and a low deployment threshold.
\paragraph{Mistral.}
The Mistral-7B-Instruct-v0.1 Large Language Model (LLM), referenced by \cite{jiang2023mistral}, is an instructionally fine-tuned variant of the Mistral-7B-v0.1 generative text model. It has been enhanced using a diverse range of publicly accessible conversational datasets.
\paragraph{Zephyr.}
The Zephyr-7B-$\beta$ \citep{tunstall2023zephyr} represents the second installment in the Zephyr series. It is a refined version of the Mistral-7B-v0.1 \citep{jiang2023mistral}, specifically enhanced through training on a combination of publicly available and synthetic datasets utilizing Direct Preference Optimization (DPO).
\paragraph{StripedHyena.}
StripedHyena \citep{stripedhyena} is the inaugural alternative model that rivals the top open-source Transformers of comparable size in both short and long-context evaluations.
\paragraph{Longchat.}
Longchat-13b-16k \citep{longchat2023} is an open-source chatbot developed through fine-tuning the llama-13b model \cite{touvron2023llama}. It utilizes conversations shared by users on ShareGPT, applying the condensing rotary embedding technique as discussed in the blog\footnote{\url{https://lmsys.org/blog/2023-06-29-longchat/}}.
\paragraph{Qwen.}
Qwen-14B-chat \citep{qwen} represents the 14-billion parameter iteration of the Qwen large language model series, abbreviated as Tongyi Qianwen, developed by Alibaba Cloud.
\paragraph{Yi.}
The Yi-34B \citep{yi2023} model represents the latest generation of open-source large language models, independently trained from scratch by 01.AI.
\paragraph{Alfred.}
Alfred-40B-1023 \citep{alfred-40b-1023} is an enhanced version of the Falcon-40B model, featuring an expanded context length capacity of 8192 tokens.
\paragraph{Beluga.}
StableBeluga-2-70B \citep{StableBelugaModels} is a variant of the Llama2 70B model \cite{touvron2023llama}, specifically fine-tuned using a dataset styled after Orca.
\paragraph{Tulu2.}
Tulu2-DPO-70B \citep{ivison2023camels} is a refined iteration of the Llama2 model \cite{touvron2023llama}, trained using a combination of publicly available, synthetic, and human datasets through the application of Direct Preference Optimization (DPO).

\section{Detailed Evaluation Results}
\begin{figure*}[!h]
  \centering
  \includegraphics[width=\textwidth]{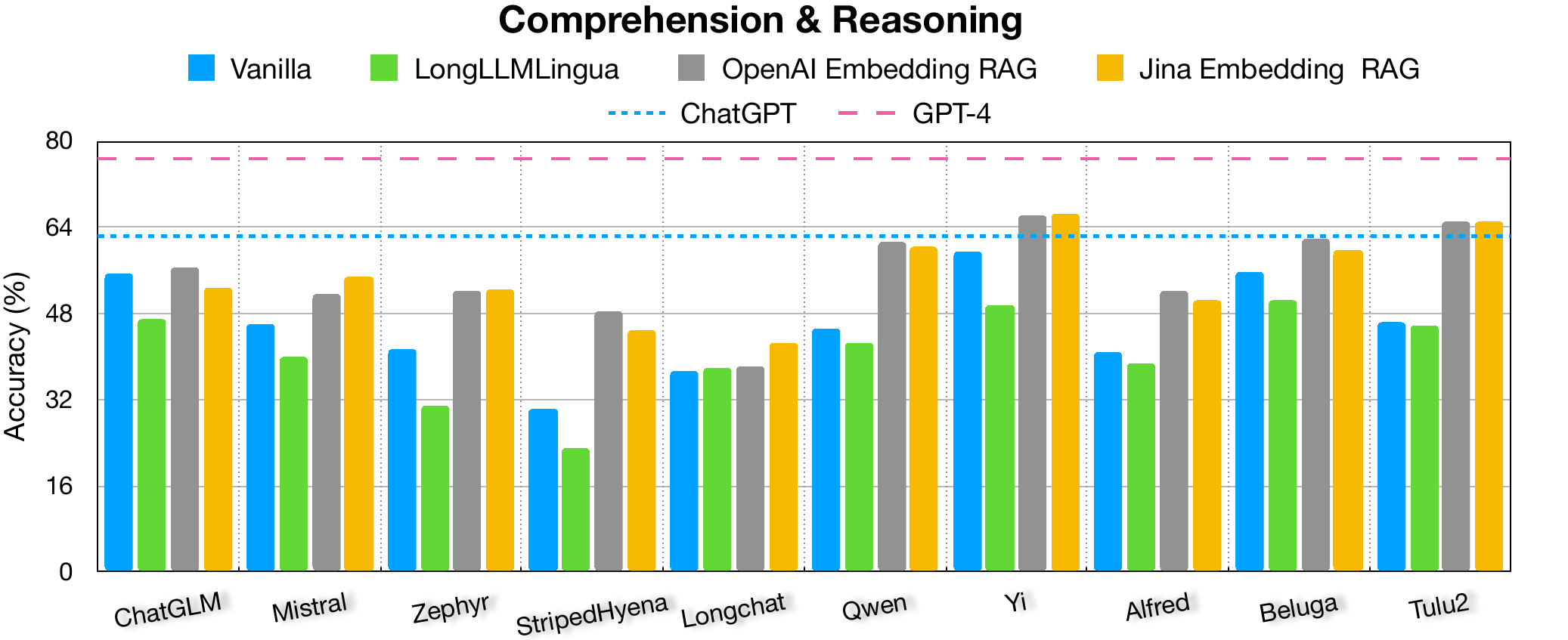}
  \caption{The performance of models on comprehension and reasoning task.}
  \label{fig:Visual_Comprehension_and_Reasoning}
\end{figure*}

\begin{figure*}[!h]
  \centering
  \includegraphics[width=\textwidth]{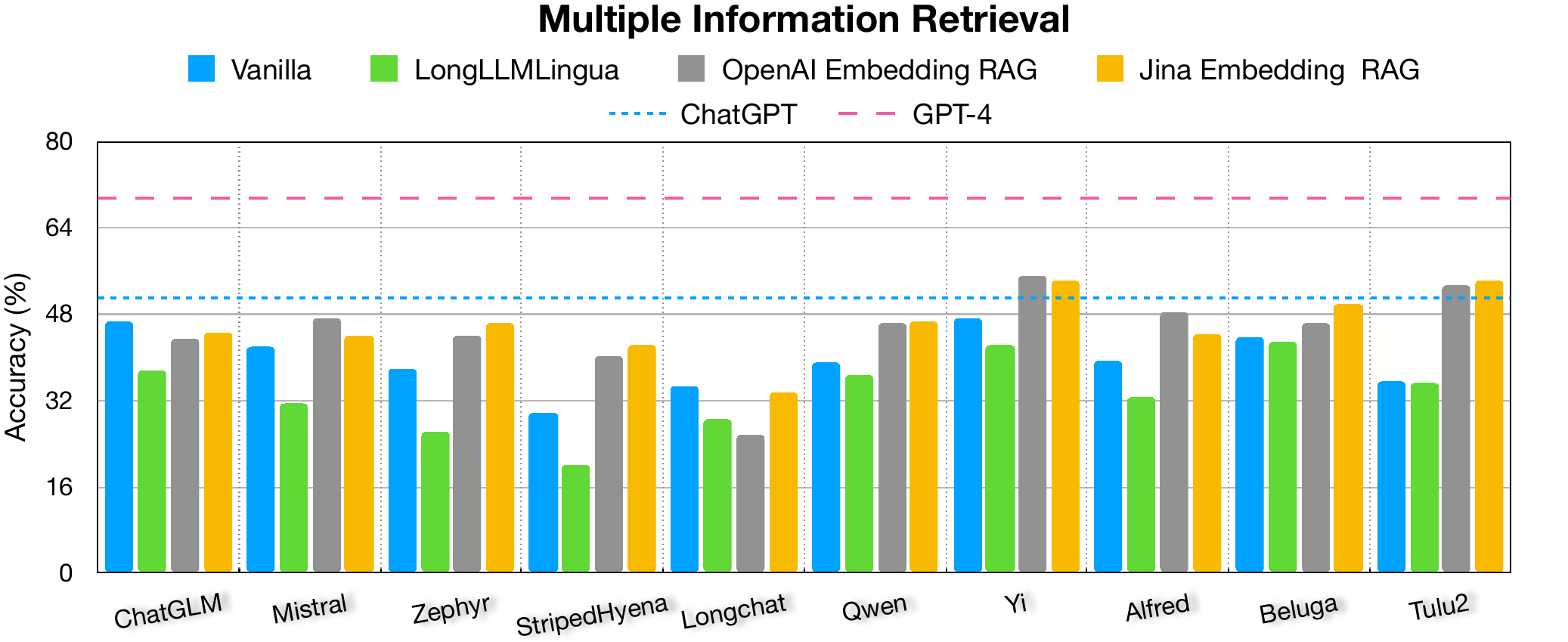}
  \caption{The performance of models on multiple information retrieval task.}
  \label{fig:Visual_Multiple_Information_Retrieval}
\end{figure*}

\begin{figure*}[!h]
  \centering
  \includegraphics[width=\textwidth]{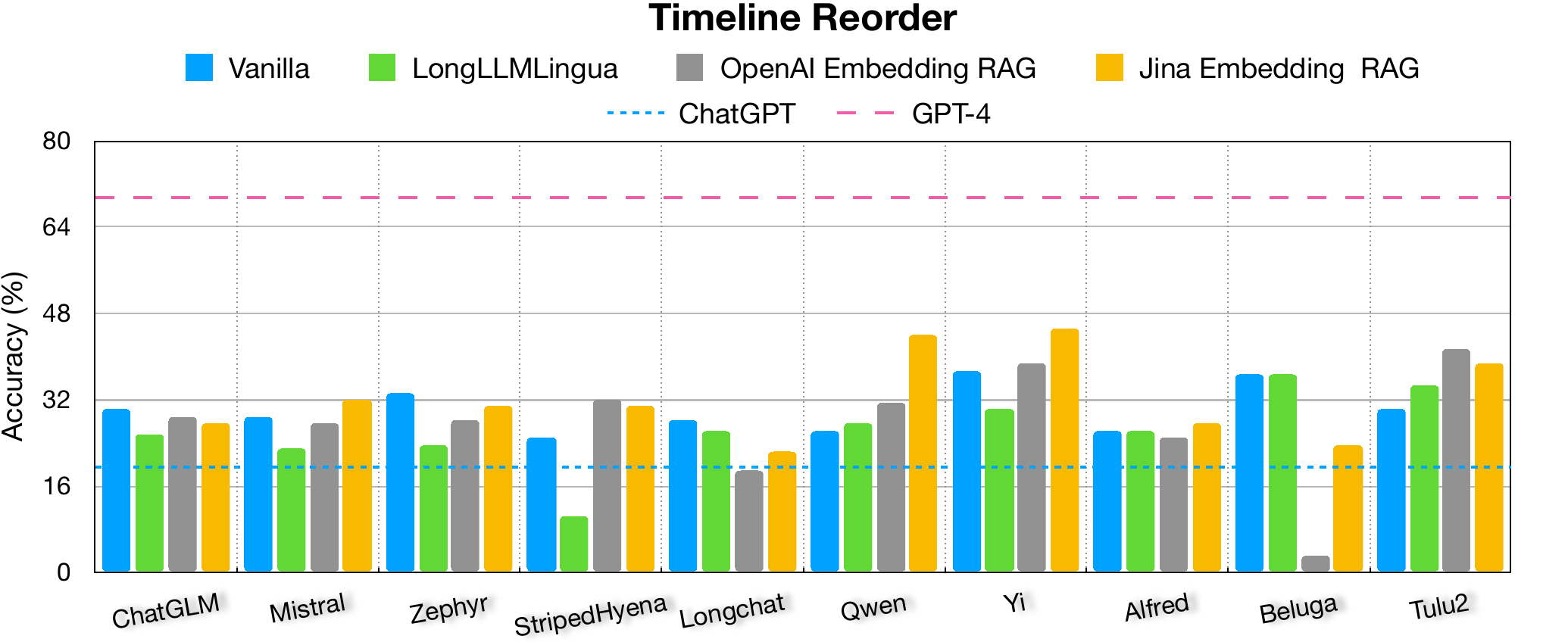}
  \caption{The performance of models on timeline reorder task.}
  \label{fig:Visual_Timeline_Reorder}
\end{figure*}

\begin{figure*}[!h]
  \centering
  \includegraphics[width=\textwidth]{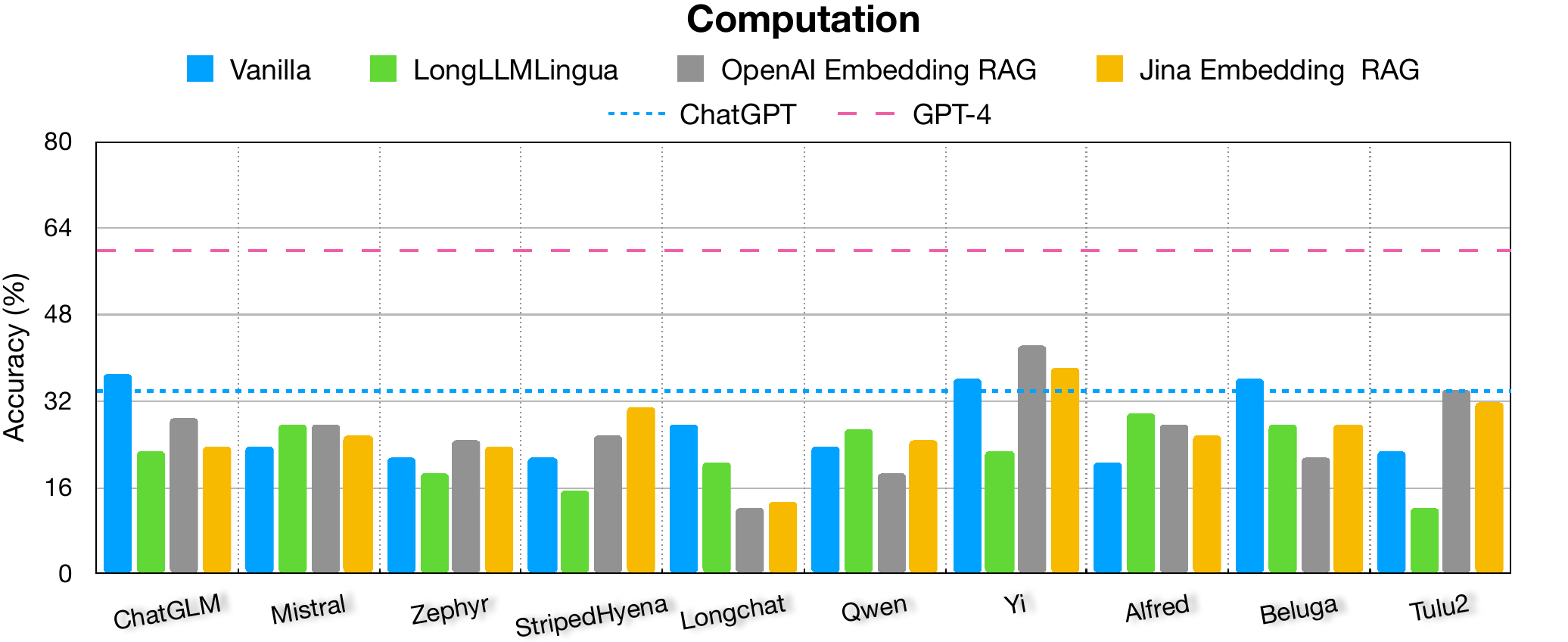}
  \caption{The performance of models on computation task.}
  \label{fig:Visual_Computation}
\end{figure*}

\begin{figure*}[!h]
  \centering
  \includegraphics[width=\textwidth]{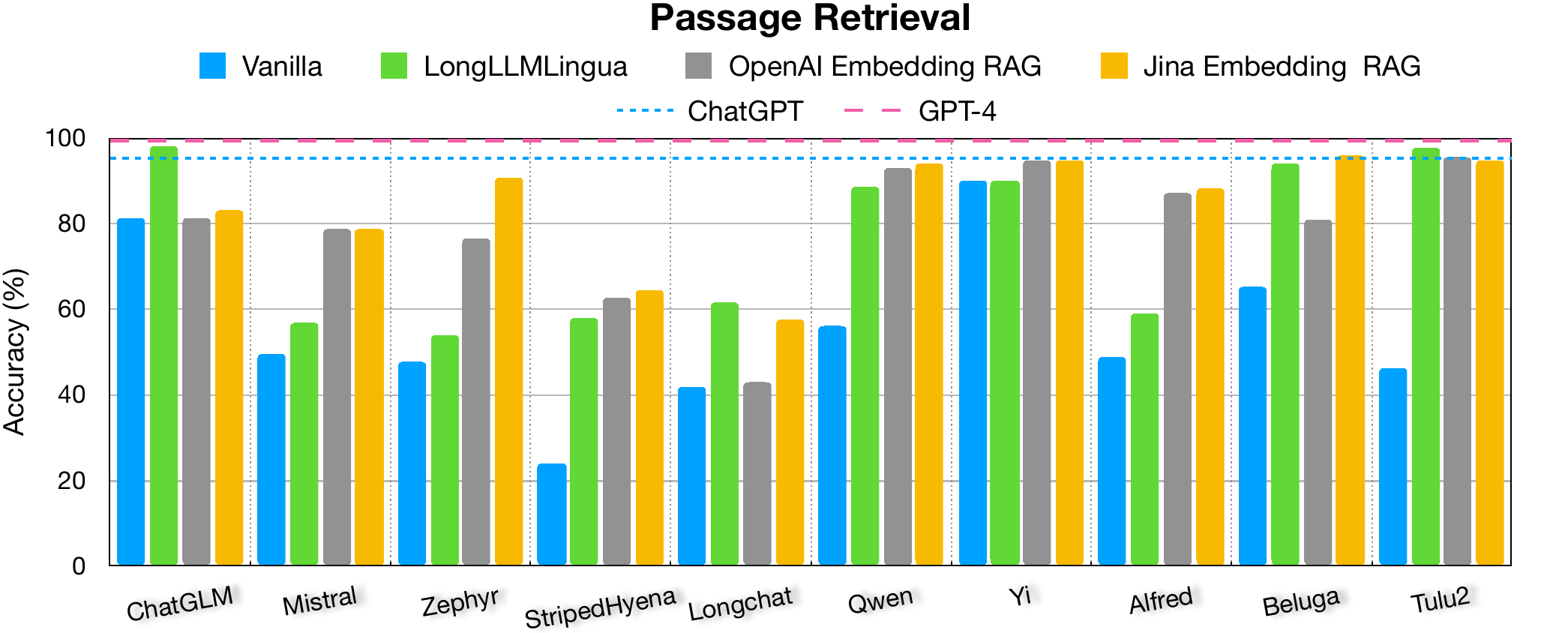}
  \caption{The performance of models on passage retrieval task.}
  \label{fig:Visual_Passage_Retrieval}
\end{figure*}

\begin{figure*}[!h]
  \centering
  \includegraphics[width=\textwidth]{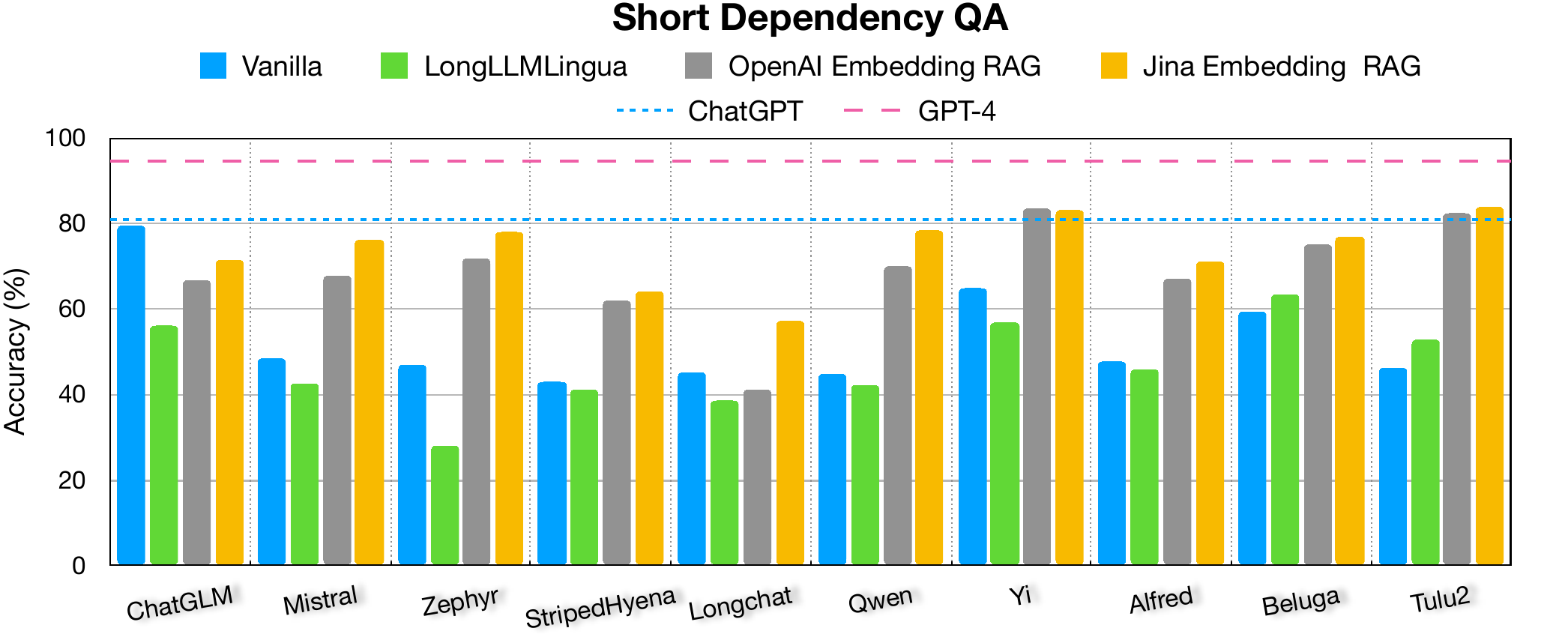}
  \caption{The performance of models on short dependency question answering task.}
  \label{fig:Visual_Short_Dependency_Question_Answering}
\end{figure*}

Table \ref{Super_Marathon_statistics} presents the detailed performance metrics of various models, utilizing distinct optimization techniques across a range of tasks.
C\&R refers to \emph{Comprehension and Reasoning} task;
MIR refers to \emph{Multiple Information Retrieval} task;
TR refers to \emph{Timeline Reorder} task;
Com. refers to \emph{Computation} task;
PR refers to \emph{Passage Retrieval} task;
SDQA refers to \emph{Short Dependency Question Answering} task;
Avg. denotes the average accuracy across all tasks.
To provide a more intuitive comparison of the effects of different optimization approaches on the long-context comprehension and reasoning capabilities of various models across different tasks, we also illustrated Figures \ref{fig:Visual_Comprehension_and_Reasoning}, \ref{fig:Visual_Multiple_Information_Retrieval}, \ref{fig:Visual_Timeline_Reorder}, \ref{fig:Visual_Computation}, \ref{fig:Visual_Passage_Retrieval}, and \ref{fig:Visual_Short_Dependency_Question_Answering}.

% \section{Template and Prompt}
% Figure \ref{fig:prompt} illustrates the prompt used in our model evaluation process.
% The system prompt is denoted in green, the provided long context in cyan, the question related to the long context in yellow, the quartet of options in blue, and the orange segment delineates the response format for the model, accompanied by an concrete example.
% The instructions and responses templates may vary across different models.
% To ensure consistency, we adjust the prompts during our evaluations to match the templates used during the models' training phases.

\section{Detailed Instruction Following Capability}
As shown in Figure \ref{fig:prompt}, we asked the models to produce results in JSON format to assess how well they follow instructions based on their output format.
Table \ref{JSON_statistics} summarizes the performance of 10 open-source large language models in this regard.
"JSON" means the output was exactly in JSON format.
"JSON-like" refers to outputs that tried to be in JSON format but included mistakes or extra text.
"Plain Text" covers outputs in other formats.
Since ChatGPT and GPT-4 can always gave results in JSON format, they're not included in Table \ref{JSON_statistics}.

\begin{table*}[!h]
  \centering
  \resizebox{\textwidth}{!}{
    \begin{tabular}{l|cccccccccc}
      \toprule
      \textbf{Type}       & \textbf{ChatGLM} & \textbf{Mistral} & \textbf{Zephyr} & \textbf{StripedHyena} & \textbf{Longchat} & \textbf{Qwen} & \textbf{Yi} & \textbf{Alfred} & \textbf{Beluga} & \textbf{Tulu2} \\
      \midrule
      \multicolumn{10}{c}{\textbf{Vanilla}}                                                                                                                                                                      \\
      \midrule
      \textbf{JSON}       & 69.48\%          & 77.22\%          & 84.51\%         & 38.43\%               & 92.29\%           & 62.29\%       & 38.95\%     & 13.27\%         & 22.48\%         & 30.72\%        \\
      \textbf{JSON-like}  & 30.52\%          & 21.18\%          & 6.86\%          & 28.95\%               & 3.99\%            & 0.72\%        & 28.56\%     & 81.96\%         & 0.33\%          & 46.47\%        \\
      \textbf{Plain Text} & 0.00\%           & 6.60\%           & 8.63\%          & 32.61\%               & 3.73\%            & 36.99\%       & 32.48\%     & 4.77\%          & 71.19\%         & 22.81\%        \\
      \midrule
      \multicolumn{10}{c}{\textbf{LongLLMLingua Compression}}                                                                                                                                                    \\
      \midrule
      \textbf{JSON}       & 94.58\%          & 68.10\%          & 63.20\%         & 50.39\%               & 93.92\%           & 90.92\%       & 48.10\%     & 13.66\%         & 29.97\%         & 35.45\%        \\
      \textbf{JSON-like}  & 5.42\%           & 22.68\%          & 9.15\%          & 12.94\%               & 2.81\%            & 0.06\%        & 26.60\%     & 85.95\%         & 0.59\%          & 43.46\%        \\
      \textbf{Plain Text} & 0.00\%           & 9.22\%           & 27.65\%         & 36.67\%               & 32.68\%           & 9.02\%        & 25.29\%     & 0.39\%          & 69.54\%         & 19.08\%        \\
      \midrule
      \multicolumn{10}{c}{\textbf{OpenAI Embedding RAG}}                                                                                                                                                         \\
      \midrule
      \textbf{JSON}       & 52.88\%          & 84.31\%          & 21.83\%         & 29.54\%               & 31.11\%           & 65.62\%       & 67.71\%     & 16.27\%         & 6.27\%          & 98.43\%        \\
      \textbf{JSON-like}  & 42.42\%          & 6.67\%           & 69.87\%         & 52.09\%               & 23.73\%           & 16.93\%       & 32.16\%     & 83.73\%         & 0.00\%          & 1.11\%         \\
      \textbf{Plain Text} & 4.71\%           & 9.02\%           & 8.30\%          & 18.37\%               & 45.16\%           & 17.45\%       & 0.13\%      & 0.00\%          & 93.73\%         & 0.46\%         \\
      \midrule
      \multicolumn{10}{c}{\textbf{Jina Embedding RAG}}                                                                                                                                                           \\
      \midrule
      \textbf{JSON}       & 86.34\%          & 83.14\%          & 17.91\%         & 24.77\%               & 32.75\%           & 63.33\%       & 65.69\%     & 0.00\%          & 8.43\%          & 97.19\%        \\
      \textbf{JSON-like}  & 9.15\%           & 6.21\%           & 73.86\%         & 56.80\%               & 21.70\%           & 17.91\%       & 34.18\%     & 100.00\%        & 0.007\%         & 2.16\%         \\
      \textbf{Plain Text} & 4.51\%           & 10.65\%          & 8.24\%          & 18.43\%               & 45.56\%           & 18.76\%       & 0.13\%      & 0.00\%          & 91.50\%         & 0.65\%         \\
      \bottomrule
    \end{tabular}
  }
  \caption{
    The evaluation results of large language models on the Marathon benchmark for instruction following.
  }
  \label{JSON_statistics}
\end{table*}

\begin{table*}
  \centering
  \resizebox{!}{0.48\textheight}{
    \begin{tabular}{l|c|c|ccccccc}
      \toprule
      \textbf{Model} & \textbf{Para.} & \textbf{CW} & \textbf{C\&R}                       & \textbf{MIR}                        & \textbf{TR}                         & \textbf{Com.}                       & \textbf{PR}                         & \textbf{SDQA}                       & \textbf{Avg.}                       \\
      \midrule
      GPT-4          & -              & 128K        & 77.03\%                             & 69.21\%                             & 69.08\%                             & 60.82\%                             & 100.00\%                            & 95.41\%                             & 78.59\%                             \\
      ChatGPT        & -              & 16K         & 62.18\%                             & 51.32\%                             & 19.74\%                             & 34.02\%                             & 95.67\%                             & 81.27\%                             & 57.37\%                             \\
      \midrule
      \multicolumn{10}{c}{\textbf{Vanilla}}                                                                                                                                                                                                                                                                                   \\
      \midrule
      ChatGLM        & 6B             & 32K         & \cellcolor[RGB]{208,255,207}55.46\% & \cellcolor[RGB]{208,255,207}46.63\% & \cellcolor[RGB]{253,255,210}30.26\% & \cellcolor[RGB]{208,255,207}37.11\% & \cellcolor[RGB]{208,255,207}81.33\% & \cellcolor[RGB]{208,255,207}79.51\% & \cellcolor[RGB]{208,255,207}55.05\% \\
      Mistral        & 7B             & 32K         & \cellcolor[RGB]{208,255,207}46.22\% & \cellcolor[RGB]{208,255,207}41.94\% & \cellcolor[RGB]{253,255,210}28.95\% & \cellcolor[RGB]{253,255,210}23.71\% & \cellcolor[RGB]{253,255,210}49.67\% & \cellcolor[RGB]{253,255,210}48.41\% & \cellcolor[RGB]{253,255,210}39.81\% \\
      Zephyr         & 7B             & 32K         & \cellcolor[RGB]{253,255,210}41.46\% & \cellcolor[RGB]{253,255,210}37.83\% & \cellcolor[RGB]{208,255,207}33.24\% & \cellcolor[RGB]{253,255,210}21.65\% & \cellcolor[RGB]{253,255,210}47.67\% & \cellcolor[RGB]{253,255,210}47.00\% & \cellcolor[RGB]{253,255,210}37.97\% \\
      StripedHyena   & 7B             & 18K         & \cellcolor[RGB]{253,255,210}30.25\% & \cellcolor[RGB]{253,255,210}29.91\% & \cellcolor[RGB]{253,255,210}25.00\% & \cellcolor[RGB]{253,255,210}21.65\% & \cellcolor[RGB]{253,255,210}24.00\% & \cellcolor[RGB]{253,255,210}43.11\% & \cellcolor[RGB]{253,255,210}28.99\% \\
      Longchat       & 13B            & 16K         & \cellcolor[RGB]{253,255,210}37.25\% & \cellcolor[RGB]{253,255,210}34.60\% & \cellcolor[RGB]{253,255,210}28.29\% & \cellcolor[RGB]{208,255,207}27.84\% & \cellcolor[RGB]{253,255,210}42.00\% & \cellcolor[RGB]{253,255,210}45.23\% & \cellcolor[RGB]{253,255,210}35.87\% \\
      Qwen           & 14B            & 8K          & \cellcolor[RGB]{253,255,210}45.38\% & \cellcolor[RGB]{253,255,210}39.00\% & \cellcolor[RGB]{253,255,210}26.32\% & \cellcolor[RGB]{253,255,210}23.71\% & \cellcolor[RGB]{208,255,207}56.33\% & \cellcolor[RGB]{253,255,210}44.88\% & \cellcolor[RGB]{253,255,210}39.27\% \\
      Yi             & 34B            & 200K        & \cellcolor[RGB]{208,255,207}59.66\% & \cellcolor[RGB]{208,255,207}47.21\% & \cellcolor[RGB]{208,255,207}37.50\% & \cellcolor[RGB]{208,255,207}36.08\% & \cellcolor[RGB]{208,255,207}90.00\% & \cellcolor[RGB]{208,255,207}65.02\% & \cellcolor[RGB]{208,255,207}55.91\% \\
      Alfred         & 40B            & 8K          & \cellcolor[RGB]{253,255,210}40.90\% & \cellcolor[RGB]{253,255,210}39.30\% & \cellcolor[RGB]{253,255,210}26.32\% & \cellcolor[RGB]{253,255,210}20.62\% & \cellcolor[RGB]{253,255,210}49.00\% & \cellcolor[RGB]{253,255,210}47.70\% & \cellcolor[RGB]{253,255,210}37.31\% \\
      Beluga         & 70B            & 4K          & \cellcolor[RGB]{208,255,207}55.74\% & \cellcolor[RGB]{208,255,207}43.70\% & \cellcolor[RGB]{208,255,207}36.84\% & \cellcolor[RGB]{208,255,207}36.08\% & \cellcolor[RGB]{208,255,207}65.33\% & \cellcolor[RGB]{208,255,207}59.36\% & \cellcolor[RGB]{208,255,207}49.51\% \\
      Tulu2          & 70B            & 8K          & \cellcolor[RGB]{208,255,207}46.50\% & \cellcolor[RGB]{253,255,210}35.48\% & \cellcolor[RGB]{253,255,210}30.26\% & \cellcolor[RGB]{253,255,210}22.68\% & \cellcolor[RGB]{253,255,210}46.33\% & \cellcolor[RGB]{253,255,210}46.29\% & \cellcolor[RGB]{253,255,210}37.92\% \\
      \midrule
      Avg.           & -              & -           & 45.88\%                             & 39.56\%                             & 30.30\%                             & 27.11\%                             & 55.17\%                             & 52.65\%                             & 41.76\%                             \\
      \midrule
      \multicolumn{10}{c}{\textbf{LongLLMLingua Compression}}                                                                                                                                                                                                                                                                 \\
      \midrule
      ChatGLM        & 6B             & 32K         & \cellcolor[RGB]{208,255,207}47.06\% & \cellcolor[RGB]{208,255,207}37.54\% & \cellcolor[RGB]{253,255,210}25.66\% & \cellcolor[RGB]{208,255,207}22.68\% & \cellcolor[RGB]{208,255,207}98.33\% & \cellcolor[RGB]{208,255,207}56.18\% & \cellcolor[RGB]{208,255,207}47.91\% \\
      Mistral        & 7B             & 32K         & \cellcolor[RGB]{253,255,210}40.06\% & \cellcolor[RGB]{253,255,210}31.38\% & \cellcolor[RGB]{253,255,210}23.03\% & \cellcolor[RGB]{208,255,207}27.84\% & \cellcolor[RGB]{253,255,210}57.00\% & \cellcolor[RGB]{253,255,210}42.76\% & \cellcolor[RGB]{253,255,210}37.01\% \\
      Zephyr         & 7B             & 32K         & \cellcolor[RGB]{253,255,210}30.81\% & \cellcolor[RGB]{253,255,210}26.39\% & \cellcolor[RGB]{253,255,210}23.68\% & \cellcolor[RGB]{253,255,210}18.56\% & \cellcolor[RGB]{253,255,210}54.00\% & \cellcolor[RGB]{253,255,210}27.92\% & \cellcolor[RGB]{253,255,210}30.23\% \\
      StripedHyena   & 7B             & 18K         & \cellcolor[RGB]{253,255,210}22.97\% & \cellcolor[RGB]{253,255,210}20.23\% & \cellcolor[RGB]{253,255,210}10.53\% & \cellcolor[RGB]{253,255,210}15.46\% & \cellcolor[RGB]{253,255,210}58.00\% & \cellcolor[RGB]{253,255,210}41.34\% & \cellcolor[RGB]{253,255,210}28.09\% \\
      Longchat       & 13B            & 16K         & \cellcolor[RGB]{253,255,210}37.82\% & \cellcolor[RGB]{253,255,210}28.74\% & \cellcolor[RGB]{253,255,210}26.32\% & \cellcolor[RGB]{253,255,210}20.62\% & \cellcolor[RGB]{253,255,210}61.67\% & \cellcolor[RGB]{253,255,210}38.52\% & \cellcolor[RGB]{253,255,210}35.61\% \\
      Qwen           & 14B            & 8K          & \cellcolor[RGB]{208,255,207}42.58\% & \cellcolor[RGB]{208,255,207}36.66\% & \cellcolor[RGB]{208,255,207}27.63\% & \cellcolor[RGB]{208,255,207}26.80\% & \cellcolor[RGB]{208,255,207}88.67\% & \cellcolor[RGB]{253,255,210}42.40\% & \cellcolor[RGB]{208,255,207}44.12\% \\
      Yi             & 34B            & 200K        & \cellcolor[RGB]{208,255,207}49.58\% & \cellcolor[RGB]{208,255,207}42.23\% & \cellcolor[RGB]{208,255,207}30.26\% & \cellcolor[RGB]{208,255,207}22.68\% & \cellcolor[RGB]{208,255,207}90.33\% & \cellcolor[RGB]{208,255,207}56.89\% & \cellcolor[RGB]{208,255,207}48.66\% \\
      Alfred         & 40B            & 8K          & \cellcolor[RGB]{253,255,210}38.94\% & \cellcolor[RGB]{253,255,210}32.84\% & \cellcolor[RGB]{253,255,210}26.32\% & \cellcolor[RGB]{208,255,207}29.90\% & \cellcolor[RGB]{253,255,210}59.00\% & \cellcolor[RGB]{253,255,210}45.94\% & \cellcolor[RGB]{253,255,210}38.82\% \\
      Beluga         & 70B            & 4K          & \cellcolor[RGB]{208,255,207}50.42\% & \cellcolor[RGB]{208,255,207}42.82\% & \cellcolor[RGB]{208,255,207}36.84\% & \cellcolor[RGB]{208,255,207}27.84\% & \cellcolor[RGB]{208,255,207}94.00\% & \cellcolor[RGB]{208,255,207}63.60\% & \cellcolor[RGB]{208,255,207}52.59\% \\
      Tulu2          & 70B            & 8K          & \cellcolor[RGB]{208,255,207}45.94\% & \cellcolor[RGB]{208,255,207}35.19\% & \cellcolor[RGB]{208,255,207}34.87\% & \cellcolor[RGB]{253,255,210}12.37\% & \cellcolor[RGB]{208,255,207}98.00\% & \cellcolor[RGB]{208,255,207}53.00\% & \cellcolor[RGB]{208,255,207}46.56\% \\
      \midrule
      Avg.           & -              & -           & 40.62\%                             & 33.40\%                             & 26.51\%                             & 22.48\%                             & 75.90\%                             & 46.86\%                             & 40.96\%                             \\
      \midrule
      \multicolumn{10}{c}{\textbf{OpenAI Embedding RAG}}                                                                                                                                                                                                                                                                      \\
      \midrule
      ChatGLM3       & 6B             & 32K         & \cellcolor[RGB]{208,255,207}56.58\% & \cellcolor[RGB]{253,255,210}43.40\% & \cellcolor[RGB]{208,255,207}28.95\% & \cellcolor[RGB]{208,255,207}28.87\% & \cellcolor[RGB]{208,255,207}81.33\% & \cellcolor[RGB]{253,255,210}66.78\% & \cellcolor[RGB]{208,255,207}50.99\% \\
      Mistral        & 7B             & 32K         & \cellcolor[RGB]{253,255,210}51.54\% & \cellcolor[RGB]{208,255,207}47.21\% & \cellcolor[RGB]{208,255,207}27.63\% & \cellcolor[RGB]{208,255,207}27.84\% & \cellcolor[RGB]{253,255,210}79.00\% & \cellcolor[RGB]{253,255,210}67.84\% & \cellcolor[RGB]{253,255,210}50.18\% \\
      Zephyr         & 7B             & 32K         & \cellcolor[RGB]{253,255,210}52.38\% & \cellcolor[RGB]{253,255,210}43.99\% & \cellcolor[RGB]{208,255,207}28.29\% & \cellcolor[RGB]{253,255,210}24.74\% & \cellcolor[RGB]{253,255,210}76.67\% & \cellcolor[RGB]{208,255,207}71.73\% & \cellcolor[RGB]{253,255,210}49.63\% \\
      StripedHyena   & 7B             & 18K         & \cellcolor[RGB]{253,255,210}48.46\% & \cellcolor[RGB]{253,255,210}40.18\% & \cellcolor[RGB]{208,255,207}32.24\% & \cellcolor[RGB]{253,255,210}25.77\% & \cellcolor[RGB]{253,255,210}62.67\% & \cellcolor[RGB]{253,255,210}62.19\% & \cellcolor[RGB]{253,255,210}45.25\% \\
      Longchat       & 13B            & 16K         & \cellcolor[RGB]{253,255,210}38.10\% & \cellcolor[RGB]{253,255,210}25.81\% & \cellcolor[RGB]{253,255,210}19.08\% & \cellcolor[RGB]{253,255,210}12.37\% & \cellcolor[RGB]{253,255,210}43.00\% & \cellcolor[RGB]{253,255,210}41.34\% & \cellcolor[RGB]{253,255,210}29.95\% \\
      Qwen           & 14B            & 8K          & \cellcolor[RGB]{208,255,207}61.34\% & \cellcolor[RGB]{208,255,207}46.33\% & \cellcolor[RGB]{208,255,207}31.58\% & \cellcolor[RGB]{253,255,210}18.56\% & \cellcolor[RGB]{208,255,207}93.00\% & \cellcolor[RGB]{208,255,207}69.96\% & \cellcolor[RGB]{208,255,207}53.46\% \\
      Yi             & 34B            & 200K        & \cellcolor[RGB]{208,255,207}66.39\% & \cellcolor[RGB]{208,255,207}55.13\% & \cellcolor[RGB]{208,255,207}38.82\% & \cellcolor[RGB]{208,255,207}42.27\% & \cellcolor[RGB]{208,255,207}95.00\% & \cellcolor[RGB]{208,255,207}83.75\% & \cellcolor[RGB]{208,255,207}63.56\% \\
      Alfred         & 40B            & 8K          & \cellcolor[RGB]{253,255,210}52.38\% & \cellcolor[RGB]{208,255,207}48.39\% & \cellcolor[RGB]{253,255,210}25.00\% & \cellcolor[RGB]{208,255,207}27.84\% & \cellcolor[RGB]{208,255,207}87.33\% & \cellcolor[RGB]{253,255,210}67.14\% & \cellcolor[RGB]{208,255,207}51.35\% \\
      Beluga         & 70B            & 4K          & \cellcolor[RGB]{208,255,207}61.90\% & \cellcolor[RGB]{208,255,207}46.33\% & \cellcolor[RGB]{253,255,210}3.28\%  & \cellcolor[RGB]{253,255,210}21.65\% & \cellcolor[RGB]{208,255,207}81.00\% & \cellcolor[RGB]{208,255,207}75.27\% & \cellcolor[RGB]{253,255,210}48.24\% \\
      Tulu2          & 70B            & 8K          & \cellcolor[RGB]{208,255,207}64.99\% & \cellcolor[RGB]{208,255,207}53.37\% & \cellcolor[RGB]{208,255,207}41.45\% & \cellcolor[RGB]{208,255,207}34.02\% & \cellcolor[RGB]{208,255,207}95.67\% & \cellcolor[RGB]{208,255,207}82.33\% & \cellcolor[RGB]{208,255,207}61.97\% \\
      \midrule
      Avg.           & -              & -           & 55.41\%                             & 45.01\%                             & 27.63\%                             & 26.39\%                             & 79.47\%                             & 68.83\%                             & 50.46\%                             \\
      \midrule
      \multicolumn{10}{c}{\textbf{Jina Embedding RAG}}                                                                                                                                                                                                                                                                        \\
      \midrule
      ChatGLM        & 6B             & 32K         & \cellcolor[RGB]{253,255,210}52.94\% & \cellcolor[RGB]{253,255,210}44.57\% & \cellcolor[RGB]{253,255,210}27.63\% & \cellcolor[RGB]{253,255,210}23.71\% & \cellcolor[RGB]{253,255,210}83.33\% & \cellcolor[RGB]{253,255,210}71.38\% & \cellcolor[RGB]{253,255,210}50.60\% \\
      Mistral        & 7B             & 32K         & \cellcolor[RGB]{253,255,210}54.90\% & \cellcolor[RGB]{253,255,210}43.99\% & \cellcolor[RGB]{253,255,210}32.24\% & \cellcolor[RGB]{253,255,210}25.75\% & \cellcolor[RGB]{253,255,210}79.00\% & \cellcolor[RGB]{208,255,207}76.33\% & \cellcolor[RGB]{253,255,210}52.04\% \\
      Zephyr         & 7B             & 32K         & \cellcolor[RGB]{253,255,210}52.66\% & \cellcolor[RGB]{208,255,207}46.33\% & \cellcolor[RGB]{253,255,210}30.92\% & \cellcolor[RGB]{253,255,210}23.71\% & \cellcolor[RGB]{208,255,207}91.00\% & \cellcolor[RGB]{208,255,207}78.09\% & \cellcolor[RGB]{208,255,207}53.79\% \\
      StripedHyena   & 7B             & 18K         & \cellcolor[RGB]{253,255,210}45.10\% & \cellcolor[RGB]{253,255,210}42.22\% & \cellcolor[RGB]{253,255,210}30.92\% & \cellcolor[RGB]{208,255,207}30.93\% & \cellcolor[RGB]{253,255,210}64.67\% & \cellcolor[RGB]{253,255,210}64.31\% & \cellcolor[RGB]{253,255,210}46.36\% \\
      Longchat       & 13B            & 16K         & \cellcolor[RGB]{253,255,210}42.58\% & \cellcolor[RGB]{253,255,210}33.43\% & \cellcolor[RGB]{253,255,210}22.37\% & \cellcolor[RGB]{253,255,210}13.40\% & \cellcolor[RGB]{253,255,210}57.67\% & \cellcolor[RGB]{253,255,210}57.24\% & \cellcolor[RGB]{253,255,210}37.78\% \\
      Qwen           & 14B            & 8K          & \cellcolor[RGB]{208,255,207}60.50\% & \cellcolor[RGB]{208,255,207}46.63\% & \cellcolor[RGB]{208,255,207}44.08\% & \cellcolor[RGB]{253,255,210}24.74\% & \cellcolor[RGB]{208,255,207}94.33\% & \cellcolor[RGB]{208,255,207}78.45\% & \cellcolor[RGB]{208,255,207}58.12\% \\
      Yi             & 34B            & 200K        & \cellcolor[RGB]{208,255,207}66.67\% & \cellcolor[RGB]{208,255,207}54.25\% & \cellcolor[RGB]{208,255,207}45.39\% & \cellcolor[RGB]{208,255,207}38.14\% & \cellcolor[RGB]{208,255,207}95.00\% & \cellcolor[RGB]{208,255,207}83.39\% & \cellcolor[RGB]{208,255,207}63.81\% \\
      Alfred         & 40B            & 8K          & \cellcolor[RGB]{253,255,210}50.42\% & \cellcolor[RGB]{253,255,210}44.28\% & \cellcolor[RGB]{253,255,210}27.63\% & \cellcolor[RGB]{253,255,210}25.77\% & \cellcolor[RGB]{208,255,207}88.33\% & \cellcolor[RGB]{253,255,210}71.02\% & \cellcolor[RGB]{253,255,210}51.24\% \\
      Beluga         & 70B            & 4K          & \cellcolor[RGB]{208,255,207}59.94\% & \cellcolor[RGB]{208,255,207}49.85\% & \cellcolor[RGB]{253,255,210}23.68\% & \cellcolor[RGB]{208,255,207}27.84\% & \cellcolor[RGB]{208,255,207}96.00\% & \cellcolor[RGB]{208,255,207}77.03\% & \cellcolor[RGB]{208,255,207}55.72\% \\
      Tulu2          & 70B            & 8K          & \cellcolor[RGB]{208,255,207}64.99\% & \cellcolor[RGB]{208,255,207}54.25\% & \cellcolor[RGB]{208,255,207}38.82\% & \cellcolor[RGB]{208,255,207}31.96\% & \cellcolor[RGB]{208,255,207}95.00\% & \cellcolor[RGB]{208,255,207}84.10\% & \cellcolor[RGB]{208,255,207}61.52\% \\
      \midrule
      Avg.           & -              & -           & 55.07\%                             & 45.98\%                             & 32.37\%                             & 26.60\%                             & 84.43\%                             & 74.13\%                             & 53.10\%                             \\
      \bottomrule
    \end{tabular}
  }
  \caption{The evaluation results of models on Marathon benchmark.}
  \label{Super_Marathon_statistics}
\end{table*}

\end{document}